\documentclass[10pt,twocolumn,letterpaper]{article}

\usepackage{cvpr}              

\definecolor{cvprblue}{rgb}{0.21,0.49,0.74}
\usepackage[pagebackref,breaklinks,colorlinks,allcolors=cvprblue]{hyperref}

\def\paperID{1279} 
\def\confName{CVPR}
\def\confYear{2025}

\usepackage{fontawesome5}
\usepackage{colortbl}
\usepackage{tabularx}
\newcommand\modelname{STiL}
\usepackage[ruled,linesnumbered]{algorithm2e}
\usepackage{array}
\usepackage{bbm}

\usepackage{lineno}
\usepackage[accsupp]{axessibility}  
\usepackage{multirow}

\newcommand{\customfootnote}[1]{
    \begingroup
    \renewcommand{\thefootnote}{}
    \footnotetext{#1}
    \addtocounter{footnote}{-1}
    \endgroup
}

\definecolor{customgreen}{HTML}{70AD47}
\definecolor{customred}{HTML}{E3201E}

\newcommand{\simf}{\mathop{\mathrm{sim}}}

\newcommand{\Avg}{\mathop{\mathrm{Avg}}}

\newcommand{\softmax}{\mathop{\mathrm{softmax}}}

\newcolumntype{P}[1]{>{\centering\arraybackslash}p{#1}}

\usepackage{environ}

\newlength{\myl}
\let\origequation=\equation
\let\origendequation=\endequation
\RenewEnviron{equation}{
  \settowidth{\myl}{$\BODY$}                       
  \origequation
  \ifdimcomp{\the\linewidth}{>}{\the\myl}
  {\ensuremath{\BODY}}                             
  {\resizebox{\linewidth}{!}{\ensuremath{\BODY}}}  
  \origendequation
}

\title{\modelname{}: Semi-supervised Tabular-Image Learning for Comprehensive Task-Relevant Information Exploration in Multimodal Classification}

\author{Siyi Du$^{1\star}$ 
\quad Xinzhe Luo$^1$
\quad Declan P. O’Regan$^2$
\quad Chen Qin$^{1\star}$\\
$^1$Department of Electrical and Electronic Engineering \& I-X, $^2$MRC Laboratory of Medical Science\\
Imperial College London, London, UK \\
{\tt\small \{s.du23, x.luo, declan.oregan, c.qin15\}@imperial.ac.uk}
}

\begin{document}

\maketitle
\begin{abstract}
    Multimodal image-tabular learning is gaining attention, yet it faces challenges due to limited labeled data. While earlier work has applied self-supervised learning (SSL) to unlabeled data, its task-agnostic nature often results in learning suboptimal features for downstream tasks. Semi-supervised learning (SemiSL), which combines labeled and unlabeled data, offers a promising solution. However, existing multimodal SemiSL methods typically focus on unimodal or modality-shared features, ignoring valuable task-relevant modality-specific information, leading to a Modality Information Gap. In this paper, we propose \modelname{}, a novel SemiSL tabular-image framework that addresses this gap by comprehensively exploring task-relevant information. \modelname{} features a new disentangled contrastive consistency module to learn cross-modal invariant representations of shared information while retaining modality-specific information via disentanglement. We also propose a novel consensus-guided pseudo-labeling strategy to generate reliable pseudo-labels based on classifier consensus, along with a new prototype-guided label smoothing technique to refine pseudo-label quality with prototype embeddings, thereby enhancing task-relevant information learning in unlabeled data. Experiments on natural and medical image datasets show that \modelname{} outperforms the state-of-the-art supervised/SSL/SemiSL image/multimodal approaches. Our code is available at \url{https://github.com/siyi-wind/STiL}.
\end{abstract}

\customfootnote{$^{\star}$Corresponding authors.}

\section{Introduction}
Multimodal deep learning (DL) involves integrating various modalities to provide a holistic understanding of subjects and is making significant advancements~\cite{baltruvsaitis2018multimodal,bayoudh2022survey}. An emerging example is image-tabular learning that combines images with structured tables, which has received increasing interest in various fields, such as marketing~\cite{he2016ups,huang2022dvm} and healthcare~\cite{acosta2022multimodal,bai2020population}. For instance, supervised image-tabular approaches~\cite{xu2016multimodal,huang2020fusion,zheng2022multi,xue2024ai} have been used to process and interpret imaging scans alongside tables (\eg, lab tests and family history), for more precise diagnosis -- similar to how clinicians assess patients in real-world settings. However, despite these achievements, such approaches often require extensive labeled training data, which are not always available, especially for classifying rare diseases.

\begin{figure}[t]
  \centering
   \includegraphics[width=1\linewidth]{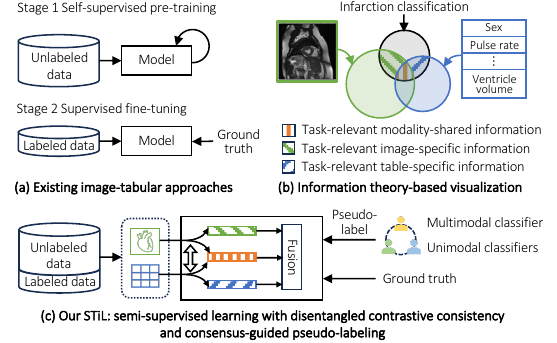}
   \caption{(a) Existing image-tabular pipelines using unlabeled data. (b) Illustration of the \emph{Information Modality Gap}: task-relevant information exists in both shared and specific features. (c) \modelname{}'s framework, which addresses this gap and effectively learns task-relevant information from labeled and unlabeled data.}
   \label{fig:challenge}
\end{figure}

To address this problem, prior image-tabular works have proposed incorporating unlabeled data through two-stage training frameworks~\cite{hager2023best,du2024tip}. They first pre-train a model on large-scale unlabeled datasets using self-supervised learning (SSL), followed by supervised fine-tuning with labeled data for downstream tasks (\cref{fig:challenge}(a)). While outperforming supervised methods, these SSL approaches still suffer significant performance drops when labeled dataset size is reduced~\cite{du2024tip,hager2023best}. This can be attributed to two main issues. First, pre-training on unlabeled data is task-agnostic, which limits the model's ability to capture information specific to downstream tasks~\cite{li2021comatch}. Second, relying solely on limited labeled data during fine-tuning increases the risk of overfitting and compromises the model's generalizability. In contrast, semi-supervised learning (SemiSL), which jointly leverages a few labeled and a large amount of unlabeled data for task-relevant information extraction, is a promising solution to overcome the above issues~\cite{van2020survey,yang2022survey}. However, to the best of our knowledge, SemiSL has not yet been explored in multimodal image-tabular learning.

Previous multimodal / multi-view SemiSL works are typically developed based on cross-modal consistency regularization~\cite{sun2020tcgm,wu2023neighbor,assefa2023audio,zhang2023multi}, co-pseudo-labeling~\cite{blum1998combining,cheng2016semi,xiong2021multiview,xu2022cross}, or a combination of both~\cite{arazo2020pseudo,wang2024knowledge,wang2024trusted}. Cross-modal consistency methods generally presume that task-relevant information primarily lies within `information intersection' across multiple modalities~\cite{sun2020tcgm}. Thus, they focus on learning cross-modal invariant (modality-shared) representations from unlabeled data by enforcing consistency constraints across modalities, \eg, contrastive constraints. On the other hand, co-pseudo-labeling methods assume that each individual modality alone can provide enough task-relevant information to train a good classifier~\cite{triguero2015self}. In these approaches, predictions from each modality-specific classifier are used to produce pseudo-labels for their counterpart modalities, which allows the model to incorporate diverse perspectives across modalities to generate reliable pseudo-labels and enhance feature extraction from unlabeled data.

However, these multimodal SemiSL approaches suffer from major bottlenecks in capturing task-relevant information in real-world applications, due to their simplified assumptions. As in~\cref{fig:challenge}(b), which shows an example in image-tabular learning, task-relevant information is not only contained in modality-shared characteristics (\eg, ventricle volume in tables and images) but is also present in modality-specific features (\eg, shape features in images, and pulse rate in tables), as with many other multimodal tasks~\cite{liang2024factorized,wang2024decoupling}. Relying solely on shared or unimodal information will thus lead to incomplete task understanding, a limitation we term as the \emph{Modality Information Gap}. Therefore, earlier cross-modal consistency methods, which focus only on shared representations, cannot fully explore task-relevant information~\cite{liang2024factorized,wang2024decoupling}. Previous co-pseudo-labeling methods also exhibit this gap, as pseudo-labels generated by unimodal classifiers or their ensembles often lack full knowledge of the task-relevant information. This can thus introduce confirmation bias~\cite{arazo2020pseudo}, \ie, accumulation of prediction errors due to unreliable pseudo-labels~\cite{yu2022semi}.

In this work, to overcome the limitations of the \emph{Modality Information Gap} and the limited availability of labeled data, we introduce \modelname{}, a novel SemiSL tabular-image framework that effectively explores both modality-shared and specific task-relevant information from labeled and unlabeled data (see~\cref{fig:challenge}(c)). Specifically, we propose a new disentangled contrastive consistency module, which uses cross-modal contrastive learning to model modality-invariant representations of shared information while preserving modality-specific features via disentanglement. We also introduce a new consensus-guided pseudo-labeling strategy that generates reliable pseudo-labels based on classifiers' consensus collaboration, to facilitate task-relevant information learning in unlabeled data while reducing confirmation bias. Inspired by the capacity of prototypes to encapsulate class-specific information~\cite{yang2023prototype,bayasi2022boosternet}, we propose a prototype-guided label smoothing strategy to further improve pseudo-label quality through prototype embeddings.

Our contributions can be summarized as follows. (1) To the best of our knowledge, this is the first work to investigate SemiSL in image-tabular learning for addressing limited labeled data. (2) We identify the \emph{Modality Information Gap} limitation in multimodal tasks and propose \modelname{}, a new SemiSL framework that fully explores task-relevant information for mitigating the gap. (3) We propose a new disentangled contrastive consistency module to learn both modality-shared and -specific information, along with novel consensus-guided pseudo-labeling and prototype-guided label smoothing strategies to improve task-relevant information learning. (4) Experiments on natural and medical datasets show \modelname{}'s remarkable outperformance over existing image/multimodal supervised/SSL/SemiSL SOTAs, particularly when labeled data is scarce.

\begin{figure*}
  \centering
\includegraphics[width=1\linewidth]{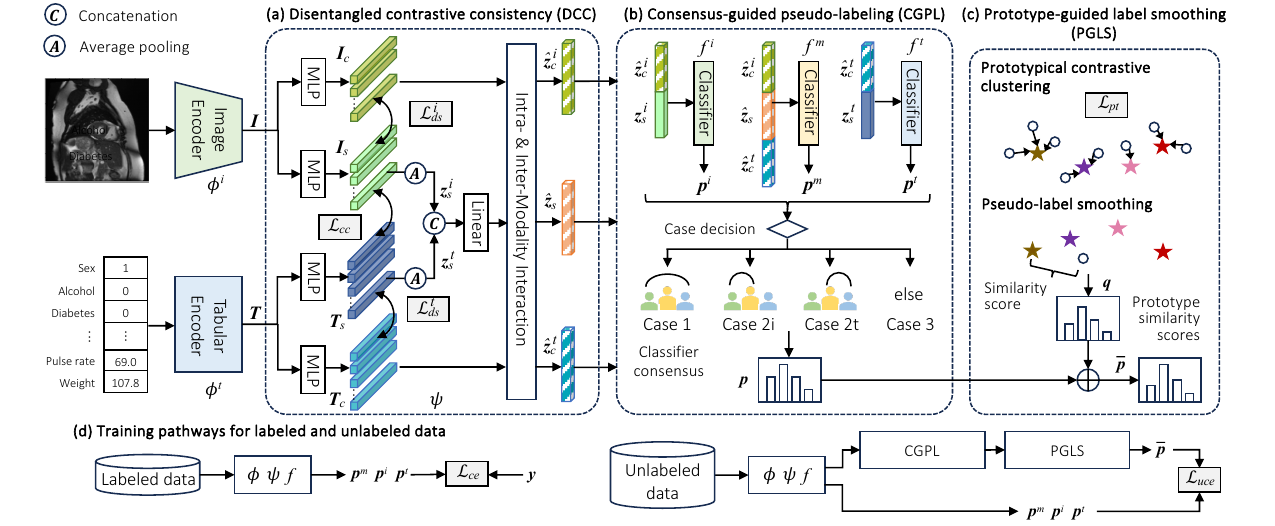}
  \caption{Overall framework of \modelname{}. \modelname{} encodes image-tabular data using $\phi$, decomposes modality-shared and -specific information through DCC $\psi$ (a), and outputs predictions via multimodal and unimodal classifiers $f$. \modelname{} generates pseudo-labels for unlabeled data using CGPL (b) and refines them with prototype similarity scores in PGLS (c). (d) Training pathways for labeled and unlabeled data.}
   \label{fig:model}
\end{figure*}

\section{Related Work}
\noindent\textbf{Semi-supervised Learning (SemiSL)} aims to reduce reliance on labeled data by exploring latent patterns from unlabeled samples~\cite{zhu2005semi,yang2022survey} and has been primarily investigated in single-modality/view settings so far. Previous works have proposed to assign pseudo-labels to unlabeled data and combine them with labeled ones to learn decision boundaries~\cite{yarowsky1995unsupervised,lee2013pseudo}. Follow-up research further improves pseudo-labeling quality through teacher-student architecture~\cite{xie2020self,pham2021meta} or uncertainty estimates~\cite{shi2018transductive,rizve2021defense}. Other studies have explored consistency regularization strategies, which enforce consistency constraints across different perturbations of the same instance, allowing the model to learn representations from unlabeled data~\cite{rasmus2015semi,sajjadi2016regularization,tarvainen2017mean,xie2020unsupervised}. Recent works have further proposed to combine these two approaches using weak-to-strong consistency regularization, where predictions from weakly augmented samples act as pseudo-labels for their strongly augmented counterparts, which has achieved promising results~\cite{zheng2022simmatch,li2021comatch,wang2022freematch,yang2023shrinking, chen2023softmatch}.

More recently, a few studies have explored SemiSL in multimodal data, proposing cross-modal consistency~\cite{sun2020tcgm,wu2023neighbor,assefa2023audio,zhang2023multi} and co-pseudo-labeling~\cite{blum1998combining,cheng2016semi,xiong2021multiview,xu2022cross,wang2024trusted}, demonstrating improved performance over unimodal models. However, these methods are typically designed for similar modalities/views, assuming no \emph{Modality Information Gap}, and are thus unable to handle dissimilar modalities (\eg, image-tabular learning).

\noindent\textbf{Multimodal Image-Tabular Learning} has received increasing attention in various domains, particularly in the medical field~\cite{huang2020fusion,polsterl2021combining,vale2021long,xue2024ai,bayasi2024continual,du2022fairdisco}. Earlier work primarily focused on designing different fusion methods~\cite{duanmu2020prediction,wolf2022daft,spasov2019parameter,zheng2022multi}, without considering the challenge of limited labeled data. MMCL~\cite{hager2023best} was the first approach to utilize self-supervised contrastive pre-training to learn representations from large-scale image-tabular pairs, followed by supervised fine-tuning on labeled data. Du \etal~\cite{du2024tip} further introduced TIP, an SSL pre-training framework to address limited and incomplete downstream task data. In contrast to these methods, \modelname{} leverages both labeled and unlabeled data jointly to enhance task-relevant information learning.

\noindent\textbf{Disentangled Representation Learning} develops models capable of decomposing specific hidden factors within data~\cite{wang2024disentangled}. A popular application is to decouple modality-shared and specific features to address challenges such as inter-modality redundancy~\cite{jia2020semi,sanchez2020learning,zhang2023prot} or missing modalities~\cite{chen2023disentangle}. Recent SSL works~\cite{liang2024factorized,wang2024decoupling} have applied this approach to mitigate modality-specific information suppression in cross-modal contrastive pre-training. However, these methods focus solely on learning separate representations for each modality, overlooking the exploration of inter-modality relations. Their task-agnostic pre-training process also limits their ability to capture task-relevant information from unlabeled data. In contrast, \modelname{} learns multimodal representations and effectively explores task-relevant information from labeled and unlabeled data.




\section{Method}
In this section, we introduce the proposed \modelname{}, the SemiSL tabular-image framework that leverages limited labeled and vast unlabeled data jointly to improve multimodal classification. We formulate the problem and outline our core idea in~\cref{sec:problem}, and discuss the details of \modelname{} in Sec. 3.2-4.

\subsection{Problem Formulation and Overall Framework}\label{sec:problem}
Let $\mathcal{X}=\lbrace(\boldsymbol{x}^i,\boldsymbol{x}^t), \boldsymbol{y}\rbrace^B$ be a batch of $B$ labeled image-tabular pairs with one-hot ground-truth labels, and $\mathcal{U}=\lbrace(\boldsymbol{u}^i,\boldsymbol{u}^t)\rbrace^{\mu B}$ be a batch of unlabeled samples, where $\mu$ is the relative size ratio between $\mathcal{X}$ and $\mathcal{U}$. Similar to~\cite{hager2023best,du2024tip}, we extract image representations $\boldsymbol{I} \in \mathbb{R}^{L^i \times D}$ using a convolutional neural network-based image encoder, and tabular representations $\boldsymbol{T} \in \mathbb{R}^{L^t \times D}$ through a transformer-based tabular encoder (\cref{fig:model}), where $L^i$ is the number of image patches and $L^t$ is the number of tabular columns.

Our goal is to enhance multimodal image-tabular classification performance by addressing the \emph{Modality Information Gap} and fully exploring task-relevant information from both labeled and unlabeled data. To achieve this, \modelname{} proposes 3 key components: (1) a disentangled contrastive consistency module (DCC, \cref{fig:model}(a), \cref{sec:DCC}), which decouples and learns comprehensive shared and specific multimodal representations; (2) a consensus-guided pseudo-labeling strategy (CGPL, \cref{fig:model}(b),~\cref{sec:CGPL}) to exploit task-relevant information in unlabeled data; and (3) a prototype-guided label smoothing strategy (PGLS, \cref{fig:model}(c),~\cref{sec:PGLS}), which further refines the pseudo-label quality.


\subsection{Disentangled Contrastive Consistency (DCC)}\label{sec:DCC}
DCC (\cref{fig:model}(a)) aims to explore comprehensive multimodal representations without relying on ground-truth supervision. To achieve this, we propose to learn modality-invariant representations of shared information by enforcing cross-modal consistency, while simultaneously decoupling modality-specific information. This will enable the model to gain a more holistic understanding of multimodal data, addressing the \emph{Modality Information Gap}. We also propose an intra- \& inter-modality interaction module to enhance both unimodal and multimodal representation learning.

\noindent\textbf{Representation Disentangling and Consistency:} To 
achieve cross-modal consistency while retaining unique information, we propose a \emph{disentanglement constraint} to disentangle shared and specific features and a \emph{shared-information consistency constraint} to ensure invariant representations for shared information between modalities. 

The \emph{disentangled constraint} aims to minimize the mutual information between shared and specific features: $MI(\boldsymbol{I}_s,\boldsymbol{I}_c)$ and $MI(\boldsymbol{T}_s,\boldsymbol{T}_c)$, where $\boldsymbol{I}_s, \boldsymbol{T}_s$ are modality-shared representations and $\boldsymbol{I}_c, \boldsymbol{T}_c$ are modality-specific representations (\cref{fig:model}(a)). Since mutual information is intractable, inspired by~\cite{zhang2023prot}, we approximate it by minimizing the CLUB loss~\cite{cheng2020club}, an upper bound of mutual information. We formulate the disentanglement losses as $\mathcal{L}_{ds}^i$ and $\mathcal{L}_{ds}^t$ (further details in Sec. A of the supplementary material).

The \emph{shared-information consistency constraint} introduces a cross-modal contrastive consistency loss $\mathcal{L}_{cc}$ based on InfoNCE~\cite{oord2018representation} for learning shared representations. Consistency is enforced on the low-dimensional representations of $\boldsymbol{I}_s$ and $\boldsymbol{T}_s$, which are derived by average pooling along the sequence dimension to obtain $\boldsymbol{z}^i_{s}$ and $ \boldsymbol{z}^t_{s}$, and then mapped to a latent space via two projection heads, $g^i$ and $g^t$. Considering all subjects in $\mathcal{X} \cup \mathcal{U}$, $L_{cc}$ is formulated as: 
\begin{align}
  \mathcal{L}_{cc} = -\frac{1}{2N}\sum_{b=1}^{N}{\left(\simf(\boldsymbol{z}_{s_b}^i,  \boldsymbol{z}_{s_b}^t)+\simf(\boldsymbol{z}_{s_b}^t, \boldsymbol{z}_{s_b}^i)\right)} &\\
  \simf(\boldsymbol{z}_{s_b}^i, \boldsymbol{z}_{s_b}^t) = \log \frac{\Psi(g^i(\boldsymbol{z}_{s_b}^i), g^t(\boldsymbol{z}_{s_b}^t))}{\sum\nolimits_{k=1}^N{\Psi(g^i(\boldsymbol{z}_{s_b}^i), g^t(\boldsymbol{z}_{s_k}^t))}}, &
  \label{eq:loss_cc} 
\end{align}
\noindent where $\Psi(\cdot, \cdot)=\exp(\cos(\cdot,\cdot)/\kappa)$, with $\kappa$ as the temperature parameter, and $N=B+\mu B$. The overall loss for DCC can be formulated as:
\begin{align}
  \mathcal{L}_{dcc} = \beta \mathcal{L}_{cc}+\gamma (\mathcal{L}_{ds}^i+\mathcal{L}_{ds}^t),
  \label{eq:loss_dcc} 
\end{align}
\noindent where $\beta$ and $\gamma$ control the weighting of the loss terms.

\noindent\textbf{Intra- \& Inter-Modality Interaction:} This module aims to exploit both intra-modality relations and the synergistic information arising from multimodal interaction~\cite{liang2024quantifying,shi2019variational}. We introduce a specialized transformer layer that incorporates self-attention on modality-specific features for extracting intra-modality dependencies, and cross-attention between shared and specific features for modeling inter-modality relations. The cross-attention (CA) is defined as:
\begin{align}
CA(\boldsymbol{Q},\boldsymbol{K},\boldsymbol{V}) = \softmax(\boldsymbol{Q}\boldsymbol{K}^T/\sqrt{d_k})\boldsymbol{V},
\end{align}
\noindent where $\boldsymbol{Q}=[\boldsymbol{z}_s]\boldsymbol{W}^Q$, $\boldsymbol{K}=[\boldsymbol{z}_s;\boldsymbol{I}_c;\boldsymbol{T}_c]\boldsymbol{W}^K$, and $\boldsymbol{V}=[\boldsymbol{z}_s;\boldsymbol{I}_c;\boldsymbol{T}_c]\boldsymbol{W}^V$. Here, $\boldsymbol{W}$ is linear transformation weights, and $\boldsymbol{z}_s$ represents shared features, defined as: $\boldsymbol{z}_s=\operatorname{Linear}(\operatorname{Concat(\boldsymbol{z}^i_s, \boldsymbol{z}^t_s)}) \in \mathbb{R}^D$. This transformer yields the enhanced shared representation $\hat{\boldsymbol{z}}_s$, along with modality-specific representations $\hat{\boldsymbol{I}}_c$ and $\hat{\boldsymbol{T}}_c$. Finally, average pooling is applied to $\hat{\boldsymbol{I}}_c$ and $\hat{\boldsymbol{T}}_c$ along the sequence dimension (the 2rd dimension), resulting in the condensed modality-specific representations $\hat{\boldsymbol{z}}^i_c$ and $\hat{\boldsymbol{z}}^t_c$ (\cref{fig:model}(a)).


\subsection{Consensus-Guided Pseudo-Labeling (CGPL)}\label{sec:CGPL}
DCC leverages unlabeled data at the feature level to learn representations in an unsupervised manner. To enable task-relevant information extraction from unlabeled data, we propose to incorporate pseudo-labels in the SemiSL process~\cite{yang2022survey}. Inspired by the success of multi-agent collaboration~\cite{minsky1988society,du2024improving,hong2023metagpt}, which shows that decisions based on multiple models are generally more robust than those from a single model, we propose CGPL (\cref{fig:model}(b)), which exploits consensus classifier collaboration to generate more reliable pseudo-labels and mitigate confirmation bias. CGPL includes two steps: consensus collaboration \& pseudo-labeling and selective classifier update.

\noindent \textbf{Consensus Collaboration \& Pseudo-Labeling:} As shown in~\cref{fig:model}(b), we construct a multimodal classifier $f^m$ using the multimodal representation, and two unimodal classifiers $f^i$ and $f^t$ using unimodal representations. To leverage classifier collaboration for pseudo-label generation, a straightforward approach is to perform an average ensemble over all classifiers. However, due to the \emph{Modality Information Gap}, unimodal classifiers lack complete task knowledge, particularly when classifying challenging samples. To alleviate this limitation, we propose a rule-based strategy for reliable pseudo-labeling, which exploits the alignment between the multimodal classifier and unimodal classifiers. Specifically, we define 4 cases: (1) Case 1: all classifiers predict the same class (agree); (2) Case 2i: $f^m$ and $f^i$ agree; (3) Case 2t: $f^m$ and $f^t$ agree; and (4) Case 3: none of the above. The pseudo-label is then determined as the average ensemble of consensus classifiers in each case (\cref{tab:CGPL}).

\noindent \textbf{Selective Classifier Update:} To reduce the risk of classifier collusion, \ie, all classifiers mistakenly agree on an incorrect class, we propose a selective updating strategy that allows classifier diversity. As shown in~\cref{tab:CGPL}, we update all classifiers in Case 1, update only the classifier with the differing predicted class in Case 2, and update either $f^i$ or $f^t$ (randomly) in Case 3. The classification loss for the unlabeled data is formulated as follows:
\begin{align}
  \mathcal{L}_{uce} = \frac{1}{\mu B}\sum_{b=1}^{\mu B}{\mathbbm{1}(\max \bar{\boldsymbol{p}}^m_b \geq \tau)\mathcal{L}(\boldsymbol{p}^m_b,\boldsymbol{p}^i_b,\boldsymbol{p}^t_b,\bar{\boldsymbol{p}}_b)},
  \label{eq:loss_uce} 
\end{align}
\noindent where $\bar{\boldsymbol{p}}^m$ and $\bar{\boldsymbol{p}}$ are refined predictions ($\boldsymbol{p}^m$ and $\boldsymbol{p}$) using PGLS, which are described in~\cref{sec:PGLS}. We retain pseudo-labels whose highest class probability is above a threshold $\tau$. The details of $\mathcal{L}(\boldsymbol{p}^m,\boldsymbol{p}^i,\boldsymbol{p}^t,\bar{\boldsymbol{p}})$ are presented in~\cref{tab:CGPL}.

\begin{table}[tb]
  \caption{Pseudo-label generation and classification loss composition for different cases. $H(\cdot,\cdot)$ denotes the cross-entropy.
  }
  \label{tab:CGPL}
  \centering
  \resizebox{0.45\textwidth}{!}{\begin{tabular}{|p{11mm}|P{24mm}|P{15mm}|P{15mm}|P{15mm}|}
    \hline
    Case & pseudo-label & \multicolumn{3}{c|}{$\mathcal{L}(\boldsymbol{p}^m,\boldsymbol{p}^i,\boldsymbol{p}^t,\bar{\boldsymbol{p}})$ (used in \cref{eq:loss_uce})} \\
    \cline{3-5}
    ~ &  $\boldsymbol{p}=$ & 
    $H(\boldsymbol{p}^m,\bar{\boldsymbol{p}})$ & $H(\boldsymbol{p}^i,\bar{\boldsymbol{p}})$ & $H(\boldsymbol{p}^t,\bar{\boldsymbol{p}})$\\
    \hline
    Case 1 & $\Avg(\boldsymbol{p}^m,\boldsymbol{p}^i,\boldsymbol{p}^t)$ & $\surd$ & $\surd$ & $\surd$\\
    \hline
    Case 2i & $\Avg(\boldsymbol{p}^m,\boldsymbol{p}^i)$ &  ~ & ~ & $\surd$ \\
    \hline
    Case 2t & $\Avg(\boldsymbol{p}^m,\boldsymbol{p}^t)$ &  ~ & $\surd$ & ~ \\
    \hline
    Case 3 & $\boldsymbol{p}^m$ & ~ & \multicolumn{2}{c|}{$\surd$ Randomly choose one} \\ 
    \hline
  \end{tabular}}
  \vspace{-7pt}
\end{table}


\subsection{Prototype-Guided Label Smoothing (PGLS)}\label{sec:PGLS}
To further enhance the reliability of pseudo-labels, we propose PGLS (\cref{fig:model}(c)), which refines pseudo-labels by incorporating feature-level label information. Unlike previous smoothing methods that rely on instance-level embeddings~\cite{li2021comatch,zheng2022simmatch}, PGLS is more efficient, as it stores only prototypes, while achieving improved performance. PGLS consists of 3 components: class prototype extraction, prototypical contrastive clustering, and pseudo-label smoothing.

\noindent \textbf{Class Prototype Extraction:} Class prototypes are defined as the mean vector of embeddings for each class. To enhance prototype reliability with limited labeled data, we propose to incorporate both labeled and confident unlabeled samples (\ie, those with $\max \bar{\boldsymbol{p}}^m \geq \tau$). Multimodal representations are projected into a low-dimensional embedding space via a projection head: $\boldsymbol{v}=h([\hat{\boldsymbol{z}}^i_c,\hat{\boldsymbol{z}}_s,\hat{\boldsymbol{z}}^t_c])$. The prototype for each class $c \in \mathcal{C}$ is then defined as:
\begin{align}
  \boldsymbol{v}_c = \frac{1}{n_c}\left(\sum^{N_l}_{y^l_j=c}{\boldsymbol{v}^l_j}+\sum^{N_u}_{\tilde{y}^u_k=c}{\mathbbm{1}(\max \bar{\boldsymbol{p}}^m_k \geq \tau)\boldsymbol{v}^u_k}\right) \label{eq:prototype}  &\\
  n_c = \sum^{N_l}_{y^l_j=c}{1}+\sum^{N_u}_{\tilde{y}^u_k=c}{\mathbbm{1}(\max \bar{\boldsymbol{p}}^m_k \geq \tau)},
  \label{eq:nc} 
\end{align}
\noindent where $N_l$ and $N_u$ are the labeled and unlabeled dataset sizes, respectively, and $\tilde{y}^u$ is the predicted class for pseudo-labels. To avoid storing instance embeddings, we maintain the sum of embeddings and $n_c$ for each class during training and compute the prototype at the end of each epoch.

\noindent \textbf{Prototypical Contrastive Clustering:} After obtaining prototype embeddings, we introduce a prototypical contrastive loss for both labeled and confident unlabeled samples, pushing them closer to their respective class prototypes and farther from other prototypes. The loss is formulated as:
\begin{equation}\label{eq:loss_pt} 
 \begin{aligned}
  \mathcal{L}_{pt} =&\ -\frac{1}{B} \sum^B_{b=1}\sum_{c \in \mathcal{C}}{\mathbbm{1}(y^l_b=c) \log \frac{\Psi(\boldsymbol{v}^l_b, \boldsymbol{v}_c)}{\sum_{c' \in \mathcal{C}} \Psi(\boldsymbol{v}^l_b, \boldsymbol{v}_{c'})}}  \\
   &- \frac{1}{\mu B}\sum^{\mu B}_{b=1} \mathbbm{1} (\max \bar{p}^m_b \geq \tau) \sum_{c \in \mathcal{C}} \mathbbm{1}(\tilde{y}^u_b=c) \log \frac{\Psi(\boldsymbol{v}^u_b, \boldsymbol{v}_c)}{\sum_{c' \in \mathcal{C}} \Psi(\boldsymbol{v}^u_b, \boldsymbol{v}_{c'})}.
   \end{aligned}
\end{equation}

\noindent \textbf{Pseudo-Label Smoothing:} Inspired by~\cite{rebuffi2017icarl}, which shows that prototype similarity (\ie, the similarity between a data sample and class prototypes) can inform classification decisions under the manifold assumption~\cite{chapelle2002cluster,roweis2000nonlinear}, we propose to smooth pseudo-labels using prototype similarity to mitigate confirmation bias (\cref{fig:model}(c)). The prototype similarity scores $\boldsymbol{q}$ are computed as: $\boldsymbol{q} = \operatorname{softmax}([\boldsymbol{v}_1,...,\boldsymbol{v}_{C}]^T\boldsymbol{v})$. The smoothed predictions are then defined as: 
\begin{align}
  \bar{\boldsymbol{p}}, \, \bar{\boldsymbol{p}}^m = r\boldsymbol{p} + (1-r) \boldsymbol{q}, \, r\boldsymbol{p}^m + (1-r) \boldsymbol{q},
  \label{eq:smooth} 
\end{align}
\noindent where $r$ controls the balance between $\boldsymbol{p}$ and $\boldsymbol{q}$. $\bar{\boldsymbol{p}}$ and $\bar{\boldsymbol{p}}^m$ are used in~\cref{eq:loss_uce},~\cref{eq:prototype},~\cref{eq:nc}, and ~\cref{eq:loss_pt}.

\noindent \textbf{Overall Loss:} The final loss of \modelname{} is as follows: 
\begin{align}
  \mathcal{L} = \alpha \mathcal{L}_{ce} + \mathcal{L}_{dcc} + \lambda_{p} \mathcal{L}_{pt} + \lambda_{u} \mathcal{L}_{uce},
  \label{eq:loss_overall} 
\end{align}
\noindent where $\mathcal{L}_{ce}=H(\boldsymbol{p}^m,\boldsymbol{y})+H(\boldsymbol{p}^i,\boldsymbol{y})+H(\boldsymbol{p}^t,\boldsymbol{y})$ is the cross-entropy loss for labeled data, and $\alpha$, $\lambda_p$, and $\lambda_{u}$ control the contributions of each respective loss term.

\noindent \textbf{Teacher-Student Framework:} To stabilize training, similar to~\cite{wang2022freematch,li2021comatch,wang2024knowledge}, we incorporate a teacher model to generate pseudo-labels and extract prototypes. This model has the same architecture as the original model (student) but is updated via exponential moving average (EMA)~\cite{he2020momentum}: $\theta' = m \theta' + (1-m) \theta$, where $m$ is the momentum coefficient. In inference, the multimodal classifier's output $\boldsymbol{p}^m$ from the student model is used as the final prediction.

\section{Experiment}

\noindent \textbf{Datasets and Evaluation Metrics:} We conduct extensive experiments on both a natural image dataset -- Data Visual Marketing (DVM)~\cite{huang2022dvm} and a medical dataset -- UK Biobank (UKBB). For UKBB, we focus on two cardiac disease classification tasks: coronary artery disease (CAD) and myocardial infarction (Infarction), using 2D short-axis cardiac magnetic resonance images (MRIs) and 75 tabular features. The dataset is split into training (26,040), validation (6,510), and test (3,617) sets. Due to low disease prevalence, we create 2 balanced training datasets for CAD (3,482) and Infarction (1,552) tasks, respectively, and evaluate the performance using the area under the curve (AUC). For DVM, we research a car model prediction task with 283 classes, using accuracy for evaluation. The DVM dataset is split into 70,565 for training, 17,642 for validation, and 88,207 for testing, with each example containing an RGB image and 17 tabular features.

\begin{table*}[tb]
  \caption{Results on DVM, CAD, and Infarction, comparing \modelname{} with supervised and SSL techniques. For SSL methods, we report results for both linear-probing (L), where the feature extractors are frozen and only the linear classifiers of the pre-trained models are tuned, and full fine-tuning (F), where all parameters are trainable. These results are indicated as (L / F).
  }
  \label{tab:SOTA_SSL}
  \centering
  \resizebox{1\textwidth}{!}{\begin{tabular}{p{36mm}|P{5mm}|P{5mm}|P{24mm}P{24mm}|P{24mm}P{24mm}|P{24mm}P{24mm}}
    \hline
    Model & \multicolumn{2}{c|}{Modality} & \multicolumn{2}{c|}{DVM Accuracy (\%) $\uparrow$} & \multicolumn{2}{c|}{CAD AUC (\%) $\uparrow$} & \multicolumn{2}{c}{Infarction AUC (\%) $\uparrow$} \\
    \cline{2-9}
    ~ & I & T & 1\% & 10\% &  1\% & 10\% &  1\% & 10\% \\
    \hline
    \multicolumn{9}{c}{(a) Supervised Methods} \\
    \hline
    ResNet-50~\cite{he2016deep} & $\surd$ & ~ & 2.85 & 32.07 & 56.93 & 50.00 & 53.30 & 55.47 \\
    DAFT~\cite{wolf2022daft} & $\surd$ & $\surd$  & 17.17 & 74.22 & 64.01 & 83.02 & 57.30 & 51.50 \\
    Interact Fuse (IF)~\cite{duanmu2020prediction} & $\surd$ & $\surd$ & 29.08 & 78.58 & 64.80 & 79.22 & 63.30 & 77.22 \\
    TIP~\cite{du2024tip} w/o SSL & $\surd$ & $\surd$ & 35.35 & 85.37 & \underline{81.13} & 83.13 & 69.07 & 58.31 \\
    \hline 
    \multicolumn{9}{c}{(b) SSL Pre-training Methods (L / F)} \\
    \hline
    SimCLR~\cite{chen2020simple}  & $\surd$ & ~ & 10.20 / 10.86 & 33.99 / 51.44 & 63.87 / 64.24 & 66.03 / 66.71 & 61.58 / 62.26 & 64.03 / 65.58 \\
    BYOL~\cite{grill2020bootstrap}  & $\surd$ & ~ & 4.90 / 5.57 & 27.32 / 47.49 & 58.22 / 59.43 & 63.38 / 63.02 & 58.32 / 58.71 & 62.55 / 62.17 \\
    SCARF~\cite{bahri2022scarf}  & ~ & $\surd$ & 38.98 / 38.53 & 61.03 / 64.47 & 66.67 / 75.76 & 81.86 / 82.43 & 65.04 / 61.76 & 77.86 / 79.92 \\
    SAINT~\cite{somepalli2021saint}  & ~ & $\surd$ & 27.98 / 1.55 & 52.60 / 83.36 & 74.75 / 78.02 & 79.21 / 83.37 & 71.39 / \underline{75.63} & 75.29 / 79.93 \\
    MMCL~\cite{hager2023best}  & $\surd$ & $\surd$ & 65.37 / 54.67 & 85.92 / 85.79 & 64.40 / 65.24 & 73.58 / 68.99 & 68.12 / 66.33 & 72.90 / 66.84 \\
    TIP~\cite{du2024tip}  & $\surd$ & $\surd$ & \underline{88.93} / 77.24 & \underline{98.75} / 98.27 & 71.17 / 77.59 & \underline{83.82} / 82.82 & 70.36 / 75.05 & \underline{80.31} / 79.71 \\
    \hline
    \rowcolor{gray!25}
    \modelname{} & $\surd$ & $\surd$ & \textbf{91.92} \textcolor{red}{\scriptsize{+2.99}} & \textbf{99.27} \textcolor{red}{\scriptsize{+0.52}} & \textbf{83.54} \textcolor{red}{\scriptsize{+2.41}} & \textbf{84.54} \textcolor{red}{\scriptsize{+0.72}} & \textbf{82.64} \textcolor{red}{\scriptsize{+7.01}} & \textbf{84.14} \textcolor{red}{\scriptsize{+3.83}} \\
    \hline
  \end{tabular}}
  \vspace{-7pt}
\end{table*}

\begin{table*}[tb]
  \caption{Results on DVM, CAD, and Infarction, comparing \modelname{} with SemiSL techniques. All methods used the same pre-trained encoders.
  }
  \label{tab:SOTA_SemiSL}
  \centering
  \resizebox{0.85\textwidth}{!}{\begin{tabular}{p{36mm}|P{5mm}|P{5mm}|P{18mm}P{18mm}|P{18mm}P{18mm}|P{18mm}P{18mm}}
    \hline
    Model & \multicolumn{2}{c|}{Modality} & \multicolumn{2}{c|}{DVM Accuracy (\%) $\uparrow$} & \multicolumn{2}{c|}{CAD AUC (\%) $\uparrow$} & \multicolumn{2}{c}{Infarction AUC (\%) $\uparrow$} \\
    \cline{2-9}
    ~ & I & T & 1\% & 10\% &  1\% & 10\% &  1\% & 10\% \\
    \hline
    CoMatch~\cite{li2021comatch} & $\surd$ & ~ & 61.12 & 86.99 & 61.09 & 71.43 & 66.48 & 71.16 \\
    SimMatch~\cite{zheng2022simmatch} & $\surd$ & ~ & 70.24 & 89.51 & 62.36 & 69.61 & 66.08 & 71.61 \\
    FreeMatch~\cite{wang2022freematch} & $\surd$ & ~ & 71.91 & 92.07 & 58.80 & 69.00 & 65.86 & 71.27 \\
    CoMatch$^{M}$ & $\surd$ & $\surd$ & 86.45 & 98.33 & 75.84 & 83.55 & 75.72 & 81.18 \\
    SimMatch$^{M}$ & $\surd$ & $\surd$ & 88.35 & 98.69 & 76.58 & \underline{84.66} & 76.07 & 80.30 \\
    FreeMatch$^{M}$ & $\surd$ & $\surd$ & 85.13 & 98.77 & 75.80 & \textbf{84.87} & 68.01 & 78.20 \\
    Co-training~\cite{blum1998combining} & $\surd$ & $\surd$ & 85.88 & \underline{98.78} & \underline{81.00} & 84.10 & \underline{79.35} & 81.05 \\
    MMatch~\cite{wang2022mmatch} & $\surd$ & $\surd$ & 87.22 & 98.76 & 79.02 & 84.35 & 78.75 & \underline{81.84} \\
    Self-KD~\cite{wang2024knowledge} & $\surd$ & $\surd$ & \underline{90.45} & 98.63 & 80.24 & 84.11 & 74.67 & 78.53 \\
    \rowcolor{gray!25}
    \modelname{} & $\surd$ & $\surd$ & \textbf{91.92} \textcolor{red}{\scriptsize{+1.47}} & \textbf{99.27} \textcolor{red}{\scriptsize{+0.49}} & \textbf{83.54} \textcolor{red}{\scriptsize{+2.54}} & 84.54 & \textbf{82.64} \textcolor{red}{\scriptsize{+3.29}} & \textbf{84.14} \textcolor{red}{\scriptsize{+2.30}} \\
    \hline
  \end{tabular}}
  \vspace{-7pt}
\end{table*}

\noindent \textbf{Implementation Details:} We used ResNet-50~\cite{he2016deep} as the image encoder and a transformer-based tabular encoder proposed by Du \etal~\cite{du2024tip}, both initialized with publicly available pre-trained weights from~\cite{du2024tip}. The hidden dimension for image and tabular representations, $\boldsymbol{I}$ and $\boldsymbol{T}$, was set to 512, and the temperature parameter $t$ was 0.1. The projection heads $g^i$ and $g^t$ for $\mathcal{L}_{cc}$ and the projection head $h$ for $\mathcal{L}_{pt}$ were 2-layer MLPs that output 128-dimensional embeddings. $f^m, f^i, f^t$ are linear classifiers. Images were resized to $128 \times 128$, and we applied augmentations to both the image and tabular data. For DVM, we set the hyperparameters as follows: $\alpha=0.2$, $\beta=3$, $\gamma=0.5$, $\lambda_p=1$, $\lambda_u=0.2$, $\tau=0.9$, $r=0.9$, $m=0.996$, $\mu=7$, $B=64$. Additional implementation details for \modelname{} and the compared models on all datasets are provided in Sec. B of supp..

\subsection{Overall Results} 

\noindent \textbf{Comparing Against Supervised/SSL SOTAs:} We conducted experiments on SOTA supervised and SSL pre-training methods. For supervised learning, we reproduced a ResNet-50 image model and 3 multimodal algorithms, \ie, DAFT~\cite{wolf2022daft}, IF~\cite{duanmu2020prediction}, and TIP~\cite{du2024tip} without pre-training weights. For SSL pre-training, we compared 2 popular image methods, SimCLR~\cite{chen2020simple} and BYOL~\cite{grill2020bootstrap}; 2 tabular methods, SCARF~\cite{bahri2022scarf} and SAINT~\cite{somepalli2021saint}; and 2 recent multimodal methods, TIP~\cite{du2024tip} and MMCL~\cite{hager2023best}. We tested SSL methods using both linear-probing, \ie, only linear classifiers are tunable, and full fine-tuning, \ie, all parameters are tunable.

\begin{table*}[tb]
  \caption{Ablation study of \modelname{}. The baseline has the same model architecture as \modelname{} but is trained using only $\mathcal{L}_{ce}$ on labeled data.
  }
  \label{tab:ablation}
  \centering
  \resizebox{0.82\textwidth}{!}{\begin{tabular}{P{13mm}|P{13mm}|P{13mm}|P{17mm}P{17mm}|P{17mm}P{17mm}|P{17mm}P{17mm}}
    \hline
    DCC & CGPL & PGLS & \multicolumn{2}{c|}{DVM Accuracy (\%) $\uparrow$} & \multicolumn{2}{c|}{CAD AUC (\%) $\uparrow$} & \multicolumn{2}{c}{Infarction AUC (\%) $\uparrow$} \\
    \cline{4-9}
    ~ & ~ & ~ & 1\% & 10\% &  1\% & 10\% &  1\% & 10\% \\
    \hline
   \multicolumn{3}{c|}{Baseline} & 82.98 & 98.04 & 75.75 & 84.13 & 73.70 & 82.52 \\
   \hline
   ~ & $\surd$ & ~ & 89.43 & \underline{98.97} & 81.26 & \underline{84.36} & 74.04 & 82.07 \\
   $\surd$ & ~ & ~ & 83.82 & 98.32 & \textbf{83.59} & 84.14 & 78.39 & 82.25 \\
   $\surd$ & $\surd$ & ~ & \underline{90.59} & 98.96 & 81.10 & 83.82 & \underline{79.53} & \underline{83.24} \\
    \rowcolor{gray!30}
    $\surd$ & $\surd$ & $\surd$ & \textbf{91.92}  & \textbf{99.27}  & \underline{83.54}  & \textbf{84.54} & \textbf{82.64} & \textbf{84.14} \\
    \hline
  \end{tabular}}
  \vspace{-10pt}
\end{table*}

\cref{tab:SOTA_SSL} shows that \modelname{} achieves the best performance in all tasks, \eg, producing a 7. 01\% higher AUC in 1\% labeled Infarction. We also observe that: (1) multimodal methods generally outperform unimodal ones, demonstrating the benefit of incorporating tabular information; (2) compared to supervised methods, SSL methods have improved performance in low-data settings but still suffer from overfitting, \eg, for MMCL and TIP on DVM, linear-probing yields better results than full fine-tuning; and (3) \modelname{} mitigates overfitting by leveraging unlabeled data during task learning, resulting in superior outcomes.

\noindent \textbf{Comparing Against SemiSL SOTAs:} We compared \modelname{} with SOTA SemiSL approaches, including 3 image methods, CoMatch~\cite{li2021comatch}, SimMatch~\cite{zheng2022simmatch}, and FreeMatch~\cite{wang2022freematch}; and 3 multimodal methods, Co-training~\cite{blum1998combining}, MMatch~\cite{wang2022mmatch} and Self-KD~\cite{wang2024knowledge}. These image methods employ strong-to-weak consistency regularization, where predictions from weakly augmented samples act as pseudo-labels for their strongly augmented versions. To evaluate the image methods in multimodal contexts, we adapted them to multimodal variants (Co/Sim/FreeMatch$^M$). Specifically, we applied strong-weak augmentations to multimodal data, concatenating unimodal features to form a multimodal representation for classification. For a fair comparison, all SemiSL methods used the same pre-trained encoders as \modelname{}.

As shown in \cref{tab:SOTA_SemiSL}, SemiSL methods outperform supervised and SSL approaches (\cref{tab:SOTA_SSL}), highlighting the benefits of leveraging unlabeled data during task learning. Adapting SemiSL image approaches (Co/Sim/FreeMatch) to multimodal settings improves their performance, but they still lag behind methods tailored for multimodal tasks. Co-training uses co-pseudo-labeling, while Self-KD relies on contrastive cross-modal consistency. However, these methods do not consider the \emph{Modality Information Gap} in multimodal tasks. Although MMatch uses a multimodal classifier for pseudo-label generation, this classifier is trained solely on a few labeled data. In contrast, \modelname{} mitigates the information gap, outperforming previous SOTAs across all tasks, \eg, 3.29\% higher AUC on 1\% labeled Infarction, which demonstrates its superior ability to fully exploit the information relevant to the task. 
Additional results for a finer grid of label percentage, different tabular encoders, and case studies are presented in Sec. C of supp..

\begin{table}[tb]
  \caption{Results of the baseline in~\cref{tab:ablation}, integrating contrastive consistency vs. disentangled contrastive consistency.
  }
  \label{tab:ablation_DCC}
  \centering
  \resizebox{0.45\textwidth}{!}{\begin{tabular}{p{46mm}|P{18mm}|P{18mm}}
    \hline
    1\% labeled data & CAD & Infarction \\
    \hline
    Baseline & 75.75 & 73.70 \\
    Baseline + $\mathcal{L}_{cc}$ & 80.95 & 71.70 \\
    Baseline + ($\mathcal{L}_{cc}$, $\mathcal{L}_{ds}^i$, $\mathcal{L}_{ds}^t$) & \textbf{83.59} & \textbf{78.39} \\
    \hline
  \end{tabular}}
  \vspace{-7pt}
\end{table}

\begin{table}[tb]
  \caption{Ablation study of DCC and CGPL on 1\% labeled data.
  }
  \label{tab:ablation_CGPL}
  \centering
  \resizebox{0.46\textwidth}{!}{\begin{tabular}{p{54mm}|P{10mm}|P{10mm}|P{10mm}}
    \hline
    1\% labeled data & DVM & CAD & Infar. \\
    \hline
    w/o intra- \& inter-modality interaction & 90.83 & 81.25 & 81.03 \\
    \hline
    w/o neither $f^{i}$ nor $f^{t}$ & 91.12 & 81.69 & 78.52 \\
    w/o consensus collaboration & 89.88 & 82.23 & 75.08 \\
    w/o selective classifier update & 91.64 & 81.17 & 80.19 \\
    w/o the EMA teacher model & 91.08 & 82.89 & 78.85 \\
    \hline
    \modelname{} & \textbf{91.92} & \textbf{83.54} & \textbf{82.64} \\
    \hline
  \end{tabular}}
  \vspace{-7pt}
\end{table}

\begin{figure*}
  \centering
  \includegraphics[width=1\linewidth]{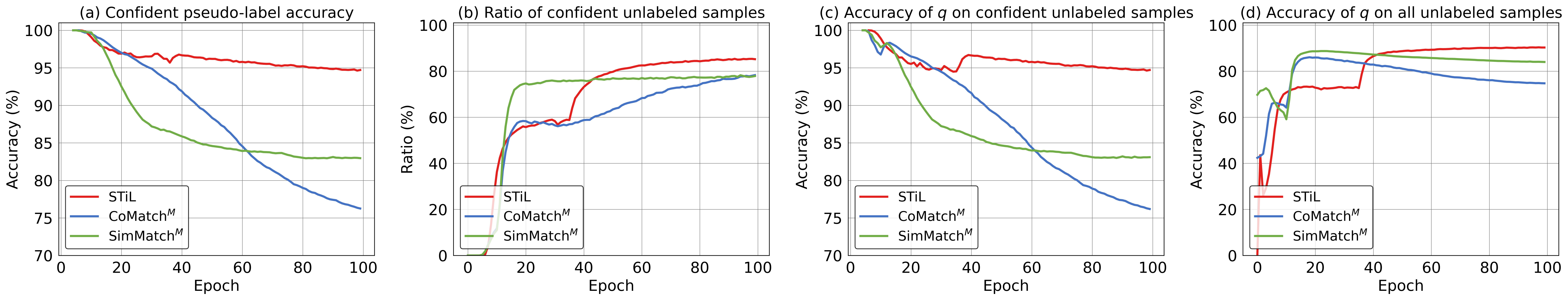}
  \caption{Plots of different methods on 1\% labeled DVM: (a) accuracy of the confident pseudo-labels, where $\max \bar{\boldsymbol{p}}^m \geq \tau$; (b) ratio of the unlabeled samples with confident pseudo-labels. (c) accuracy of the smoothness term ($\boldsymbol{q}$ in~\cref{eq:smooth}) on samples with confident pseudo-labels; and (d) accuracy of $\boldsymbol{q}$ on all unlabeled data samples.}
   \label{fig:metrics}
   \vspace{-10pt}
\end{figure*}

\subsection{Ablation Studies}

\noindent \textbf{Efficacy of Key Model Components:} We ablated the 3 proposed components of \modelname{}: DCC (\cref{sec:DCC}), CGPL (\cref{sec:CGPL}), and PGLS (\cref{sec:PGLS}). To achieve this, we established a baseline that has the same architecture as \modelname{} but is trained only with $\mathcal{L}_{ce}$ on labeled data. We then progressively incorporated each proposed component into the baseline. \cref{tab:ablation} showcases that each component improves performance, with \modelname{} -- which integrates all of them -- achieving the best results. Additionally, for the CAD task, DCC+CGPL performs worse than either DCC or CGPL individually; however, when PGLS is added, overall performance improves. This suggests that the disentanglement loss may affect pseudo-label generation, possibly due to the challenges in optimizing losses that include variational approximation~\cite{yao2018yes}. Nevertheless, adding the smoothing strategy alleviates this issue and enhances pseudo-label quality.

\noindent \textbf{Ablation Study on DCC:} We conducted ablation studies on DCC by selectively removing the modality interaction and disentanglement losses. \cref{tab:ablation_CGPL} shows that omitting our intra- \& inter-modality interaction reduces model performance. Furthermore, in~\cref{tab:ablation_DCC}, relying solely on contrastive consistency impairs model performance on Infarction, indicating that contrastive learning may overlook modality-specific information, as also noted in~\cite{liang2024factorized,wang2024decoupling}. However, our DCC effectively mitigates this issue through disentanglement and improves overall performance.

\noindent \textbf{Ablation Study on CGPL:} We designed ablation experiments to study the efficacy of the components within CGPL: (1) remove unimodal classifiers to rely solely on the multimodal classifier; (2) remove consensus collaboration, \ie, use the average ensemble of all classifiers as pseudo-labels; (3) eliminate our selective classifier update, \ie, update all classifiers simultaneously; and (4) remove the EMA teacher model. In~\cref{tab:ablation_CGPL}, these modified pseudo-labeling strategies result in a decreased performance, showing that each component is essential for pseudo-label generation. 


\begin{figure}
  \centering
  \includegraphics[width=0.9\linewidth]{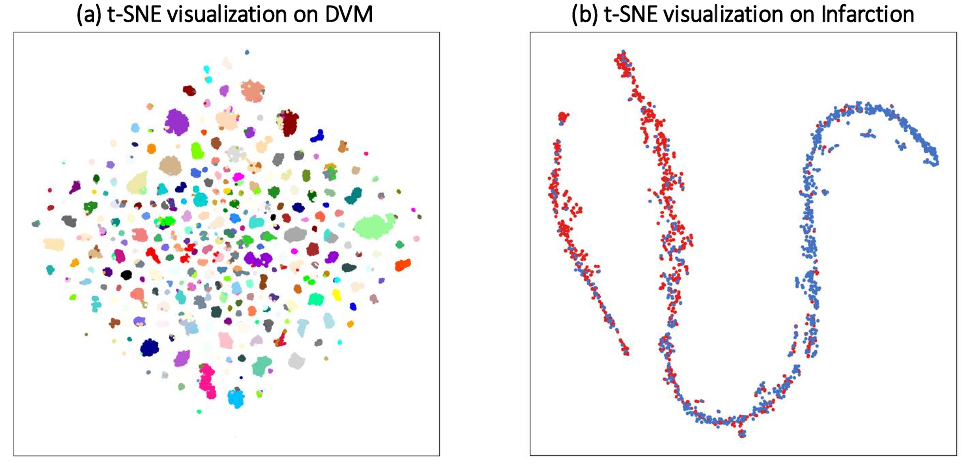}
  \caption{t-SNE visualization of the multimodal embedding $\boldsymbol{v}$ for \modelname{} trained on 1\% labeled DVM or 10\% labeled Infarction.}
   \label{fig:tsne}
   \vspace{-7pt}
\end{figure}

\begin{table}[tb]
  \caption{Number of stored embeddings for 10\% labeled data. 
  }
  \label{tab:ablation_efficiency}
  \centering
  \resizebox{0.45\textwidth}{!}{\begin{tabular}{p{21mm}|P{21mm}|P{21mm}|P{21mm}}
    \hline
    \# stored  & CoMatch$^{M}$ & SimMatch$^{M}$ & \modelname{} \\
    \hline
    DVM & 2,560 & 7,056 & \textbf{286} \\
    CAD & 2,560 & 349 & \textbf{2} \\
    \hline
  \end{tabular}}
  \vspace{-7pt}
\end{table}

\begin{table}[tb]
  \caption{Results of SemiSL methods with and w/o pre-trained weights on 1\% labeled DVM. $^{\oslash}$ denotes w/o pre-trained weights.
  }
  \label{tab:ablation_pretrain}
  \centering
  \resizebox{0.45\textwidth}{!}{\begin{tabular}{p{11mm}|P{17mm}|P{17mm}|P{17mm}|P{17mm}}
    \hline
    ~ & Self-KD$^\oslash$ & Self-KD & \modelname{}$^\oslash$ & \modelname{}  \\
    \hline
    DVM & 42.44 & 90.45 & 76.21 & \textbf{91.92} \\
    \hline
  \end{tabular}}
  \vspace{-7pt}
\end{table}

\noindent \textbf{Efficiency and Efficacy of CGPL:} We compared the efficiency of different smoothing approaches that use embedding similarities with \modelname{}. As shown in~\cref{tab:ablation_efficiency}, CoMatch$^{M}$ requires storing numerous instance embeddings, while SimMatch$^{M}$ necessitates storing the embeddings of all labeled data. However, \modelname{} only stores prototypical embeddings for each class, thus enhancing efficiency. Moreover, in \cref{fig:metrics}, our pseudo-labels and the smoothness term $\boldsymbol{q}$ demonstrate higher reliability. Finally, we used t-SNE~\cite{van2008visualizing} to visualize the multimodal embedding space. In~\cref{fig:tsne}, samples from different classes are clearly separated.

\noindent \textbf{Impact of Pre-trained Weights:} We examined the effect of using pre-trained weights. As shown in~\cref{tab:ablation_pretrain}, SemiSL approaches w/o pre-trained weights exhibit decreased performance, yet they remain significantly better than supervised methods (\cref{tab:SOTA_SSL}(a)). Furthermore, \modelname{} shows more robust performance when pre-trained weights are not available.

\begin{figure}
  \centering
  \includegraphics[width=1\linewidth]{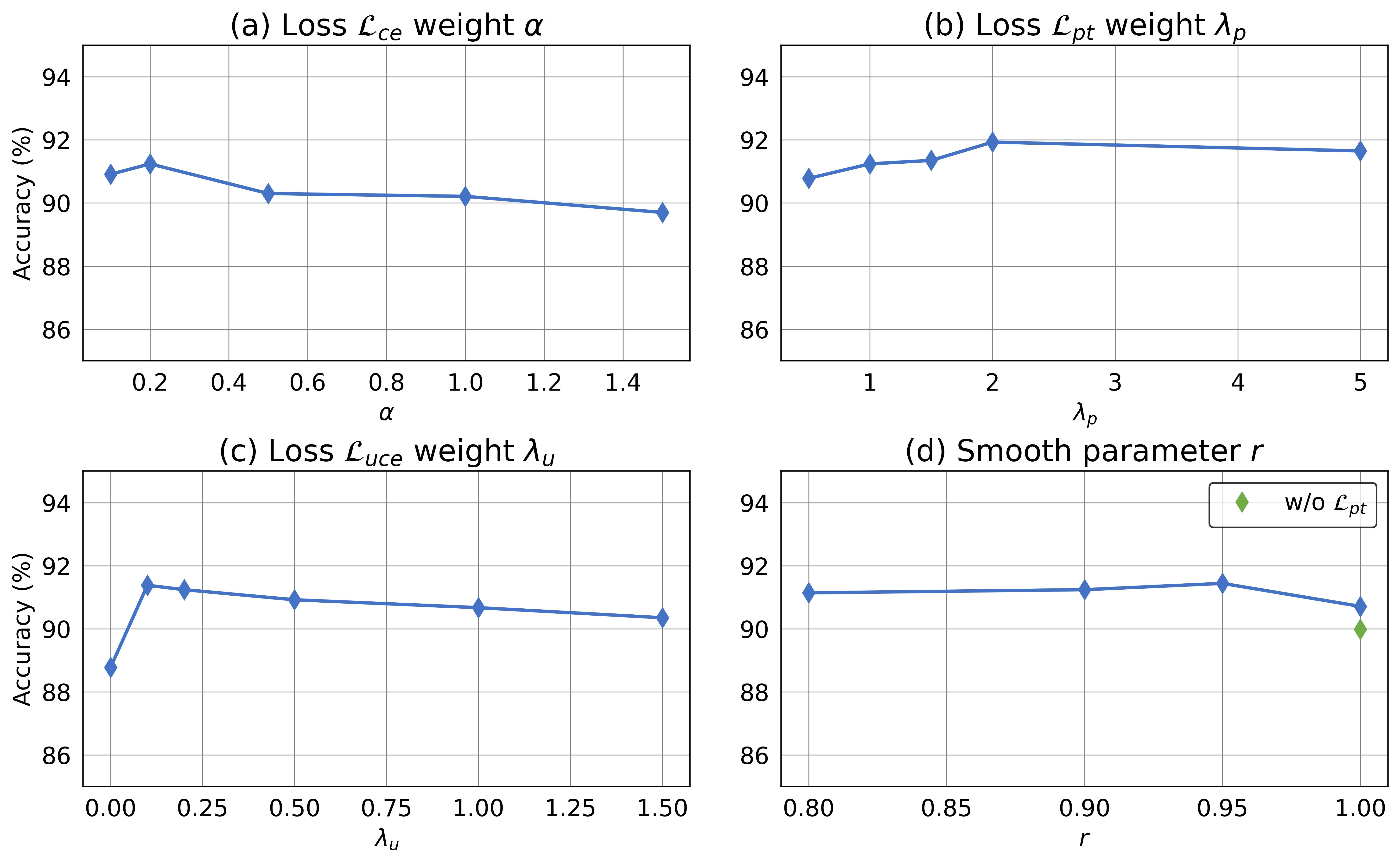}
  \caption{Results of \modelname{} on 1\% labeled DVM with varying (a) weight $\alpha$ for $\mathcal{L}_{ce}$, (b) weight $\lambda_p$ for $\mathcal{L}_{pt}$, (c) weight $\lambda_u$ for $\mathcal{L}_{uce}$, and (d) smoothness parameter $r$ in PGLS.}
   \label{fig:sensitivity}
   \vspace{-7pt}
\end{figure}

\noindent \textbf{Sensitivity Analysis:} We investigated the impact of hyper-parameters on 1\% labeled DVM. Due to the extensive number of experiments, as done in~\cite{li2021comatch}, we report the accuracy after training for 200 epochs, where the default setting of \modelname{} achieves 91.24\%. Among these parameters, $\alpha$, $\lambda_p$, and $\lambda_u$ control the contributions of $\mathcal{L}_{ce}$, $\mathcal{L}_{pt}$, and $\mathcal{L}_{uce}$, respectively (\cref{eq:loss_overall}), while $r$ is the smoothness parameter in CGPL (\cref{eq:smooth}). In~\cref{fig:sensitivity}, \modelname{}'s performance remains stable across a range of values, despite too high or too low values lead to decreased performance. This indicates \modelname{}'s relative insensitivity to hyper-parameters. Furthermore, we evaluated the effect of prototypical contrastive clustering when $r=1$ (no smoothness). In \cref{fig:sensitivity}(d), removing $\mathcal{L}_{pt}$ results in a 0.73\% accuracy drop, demonstrating that the prototype clustering also benefits the classification task.

\section{Conclusion}
In conclusion, we present the first exploration of semi-supervised learning (SemiSL) for the image-tabular domain. We propose \modelname{}, a new SemiSL framework for multimodal classification, which comprehensively explores task-relevant information from both labeled and unlabeled data, addressing the \emph{Modality Information Gap}. \modelname{} features a novel disentangled contrastive consistency module to learn modality-shared and specific representations. Additionally, we propose consensus-guided pseudo-labeling and prototype-guided label smoothing strategies to generate reliable pseudo-labels for improved task-relevant information learning. Experiments on natural and medical image datasets showed \modelname{}'s SOTA performance and the efficacy of our proposed model components. With growing interest in applying DL to image-tabular data, particularly in low-labeled data scenarios, \modelname{} offers significant potential for advancing DL deployment in this domain. Future research can generalize our approach to other multi-modalities beyond image-tabular data, such as text and video.

\section*{Acknowledgements}
This research has been conducted using the UK Biobank Resource under Application Number \emph{40616}. The MR images presented in the figures are reproduced with the kind permission of UK Biobank ©. 
{
    \small
    \bibliographystyle{ieeenat_fullname}
    \bibliography{main}
}


\clearpage
\setcounter{section}{0}
\setcounter{subsection}{0}
\setcounter{subsubsection}{0}
\setcounter{figure}{0}
\setcounter{table}{0}
\setcounter{equation}{0}

\renewcommand{\thesection}{\Alph{section}}
\renewcommand{\thesubsection}{\thesection.\arabic{subsection}}
\renewcommand{\thesubsubsection}{\thesubsection.\alph{subsubsection}}




\renewcommand{\thefigure}{S\arabic{figure}}
\renewcommand{\theequation}{S\arabic{equation}}
\renewcommand{\thetable}{S\arabic{table}}




\section{Detailed Formulations}

\noindent\textbf{Disentanglement Loss:} Our proposed disentanglement loss (Sec 3.2 of the manuscript) aims to minimize mutual information (MI) between modality-shared and modality-specific representations. As MI is intractable, we leverage an upper bound called the contrastive log-ratio upper bound (CLUB)~\cite{cheng2020club,zhang2023prot} as an MI estimator. Given sample pairs $\{(\boldsymbol{a}_j, \boldsymbol{b}_j)\}^N_j$, CLUB is defined as:
\begin{equation}
\begin{aligned}
  I_{CLUB}(\boldsymbol{a},\boldsymbol{b})& = \mathbb{E}_{p(\boldsymbol{a},\boldsymbol{b})}[\log p(\boldsymbol{b}|\boldsymbol{a})] - \mathbb{E}_{p(\boldsymbol{a})}\mathbb{E}_{p(\boldsymbol{b})}[\log p(\boldsymbol{b}|\boldsymbol{a})] \\ 
  & = \frac{1}{N}\sum_j^N\log p(\boldsymbol{b}_j|\boldsymbol{a}_j) - \frac{1}{N^2}\sum^N_{j=1}\sum^N_{k=1}\log p(\boldsymbol{b}_k|\boldsymbol{a}_j) \\
  & = \frac{1}{N^2}\sum^N_{j=1}\sum^N_{k=1}[\log p(\boldsymbol{b}_j|\boldsymbol{a}_j)-\log p(\boldsymbol{b}_k|\boldsymbol{a}_j)]
  \label{eq:CLUB},
\end{aligned}
\end{equation}
\noindent where $\log p(\boldsymbol{b}_j|\boldsymbol{a}_j)$ denotes the conditional log-likelihood of a positive sample pair $(\boldsymbol{a}_j,\boldsymbol{b}_j)$, and $\{\log p(\boldsymbol{b}_k|\boldsymbol{a}_j)\}_{j \neq k}$ is the conditional log-likelihood of a negative sample pair $(\boldsymbol{a}_j,\boldsymbol{b}_k)$. However, as \modelname{} obtains modality-shared and modality-specific representations simultaneously during training, the exact conditional distribution between these two representations is unavailable. To address this limitation, similar to~\cite{cheng2020club,zhang2023prot}, we leverage a variational distribution $q_{\theta}(\boldsymbol{b}|\boldsymbol{a})$ (an MLP layer with parameter $\theta$) to approximate $p(\boldsymbol{b}|\boldsymbol{a})$. This leads to a variational CLUB (vCLUB), formulated as:
\begin{equation}
\begin{aligned}
  I_{vCLUB}(\boldsymbol{a},\boldsymbol{b})& = \mathbb{E}_{p(\boldsymbol{a},\boldsymbol{b})}[\log q_{\theta}(\boldsymbol{b}|\boldsymbol{a})] - \mathbb{E}_{p(\boldsymbol{a})}\mathbb{E}_{p(\boldsymbol{b})}[\log q_{\theta}(\boldsymbol{b}|\boldsymbol{a})] \\ 
  & = \frac{1}{N}\sum_j^N\log q_{\theta}(\boldsymbol{b}_j|\boldsymbol{a}_j) - \frac{1}{N^2}\sum^N_{j=1}\sum^N_{k=1}\log q_{\theta}(\boldsymbol{b}_k|\boldsymbol{a}_j) \\
  & = \frac{1}{N^2}\sum^N_{j=1}\sum^N_{k=1}[\log q_{\theta}(\boldsymbol{b}_j|\boldsymbol{a}_j)-\log q_{\theta}(\boldsymbol{b}_k|\boldsymbol{a}_j)].
\end{aligned}
\end{equation}

To enforce $q_{\theta}(\boldsymbol{b}|\boldsymbol{a})$ align closely with $p(\boldsymbol{b}|\boldsymbol{a})$, we maximize the following log-likelihood:
\begin{align} 
\mathcal{L}_{q_{\theta}}(\boldsymbol{a},\boldsymbol{b})=\frac{1}{N}\sum^N_j \log q_{\theta}(\boldsymbol{b}_j|\boldsymbol{a}_j).
\end{align}
Finally, our disentanglement losses $\mathcal{L}_{ds}^i$ and $\mathcal{L}_{ds}^t$ can be formulated as:
\begin{align} 
\mathcal{L}_{ds}^i = I_{vCLUB}(\boldsymbol{z}_c^i, \boldsymbol{z}_s^i)-\mathcal{L}_{q_{\theta}}(\boldsymbol{z}_c^i, \boldsymbol{z}_s^i) \\
\mathcal{L}_{ds}^t = I_{vCLUB}(\boldsymbol{z}_c^t, \boldsymbol{z}_s^t)-\mathcal{L}_{q_{\theta}}(\boldsymbol{z}_c^t, \boldsymbol{z}_s^t),
\end{align}
\noindent where $\boldsymbol{z}^i_c$ and $\boldsymbol{z}^t_c$ are modality-specific representations and $\boldsymbol{z}^i_s$ and $\boldsymbol{z}^t_s$ are modality-shared representations. These two losses are used in Eq. (3) of the manuscript.

\begin{table}[tb]
  \caption{Definitions of symbols used for \modelname{}'s hyper-parameters.
  }
  \label{tab:symbol}
  \centering
  \resizebox{0.46\textwidth}{!}{\begin{tabular}{p{4mm}|p{102mm}}
    \hline
    ~ & Description \\
    \hline
    $B$ & Batch size of labeled data \\
    $\mu$ & Relative size ratio between labeled and unlabeled batches \\
    $\alpha$ & Weighting coefficient controlling the labeled cross-entropy loss $\mathcal{L}_{ce}$ \\
    $\beta$ & Weighting coefficient controlling the contrastive consistency loss $\mathcal{L}_{cc}$ \\
    $\gamma$ & Weighting coefficient controlling the disentanglement losses $\mathcal{L}_{ds}^i$ and $\mathcal{L}_{ds}^t$ \\
    $\lambda_p$ & Weighting coefficient controlling the prototypical contrastive loss $\mathcal{L}_{pt}$ \\
    $\lambda_u$ & Weighting coefficient controlling the unlabeled cross-entropy loss $\mathcal{L}_{uce}$ \\
    $\tau$ & Threshold for defining confident pseudo-labels \\
    $r$ & Smoothness Weighting coefficient in PGLS\\
    $m$ & Momentum coefficient for EMA \\
    $\kappa$ & Temperature parameter \\
    \hline
  \end{tabular}}
\end{table}

\begin{table}[tb]
  \caption{Hyper-parameter settings for \modelname{}.
  }
  \label{tab:hyperparameter}
  \centering
  \resizebox{0.46\textwidth}{!}{\begin{tabular}{p{15mm}|P{5mm}P{5mm}P{5mm}P{5mm}P{5mm}P{5mm}P{5mm}P{6mm}P{6mm}P{8mm}P{5mm}}
    \hline
    Task & $B$ & $\mu$ & $\alpha$ & $\beta$ & $\gamma$ & $\lambda_p$ & $\lambda_u$ & $\tau$ & $r$ & $m$ & $\kappa$\\
    \hline
    DVM & 64 & 7 & 0.2 & 3 & 0.5 & 1 & 0.2 & 0.9 & 0.9 & 0.996 & 0.1 \\
    \hline
    1\% CAD & \multirow{2}{*}{32} & \multirow{2}{*}{7} & \multirow{2}{*}{0.2} & \multirow{2}{*}{0.5} & \multirow{2}{*}{5} & \multirow{2}{*}{0.5} & \multirow{2}{*}{5} & 0.85 & \multirow{2}{*}{0.95} & \multirow{2}{*}{0.4} & \multirow{2}{*}{0.1}\\
    10\% CAD & ~ & ~ & ~ & ~ & ~ & ~ & ~ & 0.8 & ~ & ~ & ~\\
    \hline 
    1\% Infa. & \multirow{2}{*}{32} & \multirow{2}{*}{7} & \multirow{2}{*}{0.2} & \multirow{2}{*}{1} & \multirow{2}{*}{1} & \multirow{2}{*}{0.5} & \multirow{2}{*}{2} & 0.85 & \multirow{2}{*}{0.95} & \multirow{2}{*}{0.4} & \multirow{2}{*}{0.1}\\
    10\% Infa. & ~ & ~ & ~ & ~ & ~ & ~ & ~ & 0.8 & ~ & ~ & ~\\
    \hline
  \end{tabular}}
\end{table}

\section{Implementation Details}
\noindent \textbf{Datasets:} The UK Biobank (UKBB) dataset~\cite{sudlow2015uk} consists of magnetic resonance images (MRIs) and tabular data related to cardiac diseases. Following prior work~\cite{du2024tip,hager2023best}, we used mid-ventricle slices from cardiac MRIs in three
time phases, \ie, end-systolic (ES) frame, end-diastolic (ED) frame, and an intermediate time
frame between ED and ES. In addition, we employed 75 disease-related tabular features, including 26 categorical features (\eg, alcohol drinker status) and 49 continuous
features (\eg, average heart rate). The DVM dataset~\cite{huang2022dvm} includes 2D RGB car images along with tabular data describing the characteristics of the car. As done in~\cite{du2024tip,hager2023best}, we employed 17 tabular features, including 4 categorical features (\eg, color), and 13 continuous features (\eg, width). Detailed benchmark information can be found in the supplementary material of~\cite{du2024tip}.

\begin{table*}[tb]
  \caption{Number of parameters and learning rates for DVM, CAD, and Infarction across different algorithms. We provide the number of parameters used during both training and testing. For SSL methods, the learning rates are reported for both linear-probing (L), where the feature extractors are frozen and only the linear classifiers of the pre-trained models are tuned, and full fine-tuning (F), where all parameters are trainable. Learning rates are indicated as (L / F). “M” denotes millions, and “1e-3” represents $1 \times 10^{-3}$.
  }
  \label{tab:learning_rate}
  \centering
  \resizebox{1\textwidth}{!}{\begin{tabular}{p{34mm}|P{5mm}|P{5mm}|P{40mm}P{30mm}|P{40mm}P{30mm}}
    \hline
    Model & \multicolumn{2}{c|}{Modality} & \multicolumn{2}{c|}{DVM} & \multicolumn{2}{c}{CAD \& Infarction} \\
    \cline{2-7}
    ~ & I & T & \#Params (train/test) & learning rate &  \#Params (train/test) & learning rate  \\
    \hline
    \multicolumn{7}{c}{(a) Supervised Methods} \\
    \hline
    ResNet-50~\cite{he2016deep} & $\surd$ & ~ & 24.1M / 24.1M & 3e-4 & 23.5M / 23.5M & 1e-3  \\
    DAFT~\cite{wolf2022daft} & $\surd$ & $\surd$  & 26.0M / 26.0M & 3e-4 & 25.4M / 25.4M & 3e-3  \\
    IF~\cite{duanmu2020prediction} & $\surd$ & $\surd$ & 26.9M / 26.9M & 3e-4 & 26.3M / 26.3M & 3e-3  \\
    TIP~\cite{du2024tip} w/o SSL & $\surd$ & $\surd$ & 54.2M / 54.2M & 3e-4 & 54.1M / 54.1M & 3e-3  \\
    \hline 
    \multicolumn{7}{c}{(b) SSL Pre-training Methods (L / F)} \\
    \hline
    SimCLR~\cite{chen2020simple}  & $\surd$ & ~ & 28.0M / 24.1M  & 1e-3 / 1e-4  & 28.0M / 23.5M  & 1e-3 / 1e-3   \\
    BYOL~\cite{grill2020bootstrap}  & $\surd$ & ~ & 70.1M / 24.1M & 1e-3 / 1e-4 & 70.1M / 23.5M  & 1e-3 / 1e-4   \\
    SCARF~\cite{bahri2022scarf}  & ~ & $\surd$ & 0.6M / 0.4M  & 1e-4 / 1e-4  & 0.7M / 0.3M  & 1e-3 / 1e-3   \\
    SAINT~\cite{somepalli2021saint}  & ~ & $\surd$ & 6.5M / 6.5M  & 1e-4 / 1e-5  & 99.1M / 99.1M  & 1e-3 / 1e-5  \\
    MMCL~\cite{hager2023best}  & $\surd$ & $\surd$ & 36.8M / 24.1M  & 1e-3 / 1e-3  & 36.9M / 23.5M  & 1e-3 / 1e-3   \\
    TIP~\cite{du2024tip}  & $\surd$ & $\surd$ & 58.8M / 54.2M  & 1e-4 / 1e-4  & 58.9M / 54.1M  & 1e-3 / 1e-4  \\
    \hline
    \multicolumn{7}{c}{(c) SemiSL Methods} \\
    \hline
    CoMatch~\cite{li2021comatch} & $\surd$ & ~ & 28.6M / 24.1M & 1e-4 & 28.0M / 23.5M & 1e-3   \\
    SimMatch~\cite{zheng2022simmatch} & $\surd$ & ~ & 28.6M / 24.1M & 1e-4 & 28.0M / 23.5M  & 1e-3   \\
    FreeMatch~\cite{wang2022freematch} & $\surd$ & ~ & 28.6M / 24.1M & 1e-4 & 28.0M / 23.5M  & 1e-3   \\
    CoMatch$^{M}$ & $\surd$ & $\surd$ & 38.1M / 37.5M & 1e-4 & 37.8M / 37.3M & 1e-3  \\
    SimMatch$^{M}$ & $\surd$ & $\surd$ & 38.1M / 37.5M & 1e-4 & 37.8M / 37.3M & 1e-3  \\
    FreeMatch$^{M}$ & $\surd$ & $\surd$ & 38.1M / 37.5M & 1e-4 & 37.8M / 37.3M & 1e-3  \\
    Co-training~\cite{blum1998combining} & $\surd$ & $\surd$ & 38.1M /38.1M & 1e-4 & 37.4M / 37.4M  & 1e-3  \\
    MMatch~\cite{wang2022mmatch} & $\surd$ & $\surd$ & 38.1M / 38.1M & 1e-4 & 37.4M / 37.4M & 1e-3  \\
    Self-KD~\cite{wang2024knowledge} & $\surd$ & $\surd$ & 44.0M / 44.0M & 1e-4 & 43.6M / 43.6M & 1e-3  \\
    \modelname{} & $\surd$ & $\surd$ & 46.7M / 43.0M  & 1e-4  & 46.2M / 42.0M  & 1e-3 \\
    \hline
  \end{tabular}}
\end{table*}

To construct a training dataset with 10\% labeled samples, we randomly sampled 10\% of the labeled instances from each class, ensuring that the class distribution remains consistent with the original training dataset. A similar procedure is followed when creating the 1\% labeled dataset. We adopted the same data augmentation technique as described in~\cite{hager2023best,du2024tip}. For image data, we employed random scaling, rotation, shifting, flipping, Gaussian noise, as well as brightness, saturation, and contrastive changes, followed by resizing the images to $128 \times 128$. For tabular data, categorical values (\eg, 'yes', 'no', and 'blue') were converted into ordinal numbers, while continuous (numerous) values were standardized using z-score normalization. To enhance data diversity, we randomly replaced 30\% of the tabular values for each subject with random values from the respective columns. Note that tabular SSL models (SCARF~\cite{bahri2022scarf} and SAINT~\cite{somepalli2021saint}) implement their own tabular augmentation strategies. The hyper-parameters and training configurations for the supervised and SSL models were consistent with those used in~\cite{hager2023best,du2024tip}. The batch size was set to 512 for DVM and 256 for both CAD and Infarction. The hyper-parameter settings for the proposed \modelname{} and SemiSL algorithms are detailed below.

\noindent\textbf{The Proposed \modelname{}:} We used ResNet-50 as the image encoder and a transformer-based tabular encoder proposed by Du \etal~\cite{du2024tip}, both initialized with publicly available pre-trained weights from~\cite{du2024tip}. The tabular encoder consists of 4 transformer layers, each with 8 attention heads and a hidden dimension of 512. For a fair comparison, all SemiSL methods used the same pre-trained encoders. Details of \modelname{}'s hyper-parameters and their configurations are provided in~\cref{tab:symbol}
 and~\cref{tab:hyperparameter}, respectively. Based on validation performance, we set the starting pseudo-labeling epoch to 25 for 10\% labeled DVM, 35 for 1\% labeled DVM, and 8 for both CAD and Infarction. The GFLOPS for STiL is 3.63.

\noindent\textbf{CoMatch~\cite{li2021comatch}:} This framework relies on strong-to-weak consistency regularization and contrastive learning. It refines pseudo-labels by incorporating information from nearby samples in the embedding space, and then uses these pseudo-labels to regulate the structure of embeddings via graph-based contrastive learning. Following the original paper, we set the weight factors for unlabeled classification loss and contrastive loss, $\lambda_{cls}$ and $\lambda_{ctr}$, to 10. The smoothness parameter $\alpha$ was set to 0.9, the embedding memory bank size $K$ to 2,560, the temperature parameter to 0.1, and the EMA momentum to 0.996. The batch sizes for the labeled and unlabeled data were the same as those used in \modelname{}. Additionally, based on validation performance, we set the thresholds for strong-to-weak consistency and graph-based contrastive learning as follows: $\tau=0.8$
and $T=0.6$ for DVM, and $\tau=0.6$ and $T=0.3$ for both CAD and Infarction. The starting pseudo-labeling epoch was 10 for DVM and 8 for CAD and Infarction.

\noindent\textbf{CoMatch$^{M}$:} This model is an extension of CoMatch to the multimodal image-tabular setting. Its hyper-parameters were the same as those in CoMatch. Based on validation performance, we set the thresholds for strong-to-weak consistency and graph-based contrastive learning as follows: $\tau=0.9$
and $T=0.8$ for DVM, and $\tau=0.85$ and $T=0.7$ for CAD and Infarction.

\begin{figure*}
  \centering
  \includegraphics[width=0.75\linewidth]{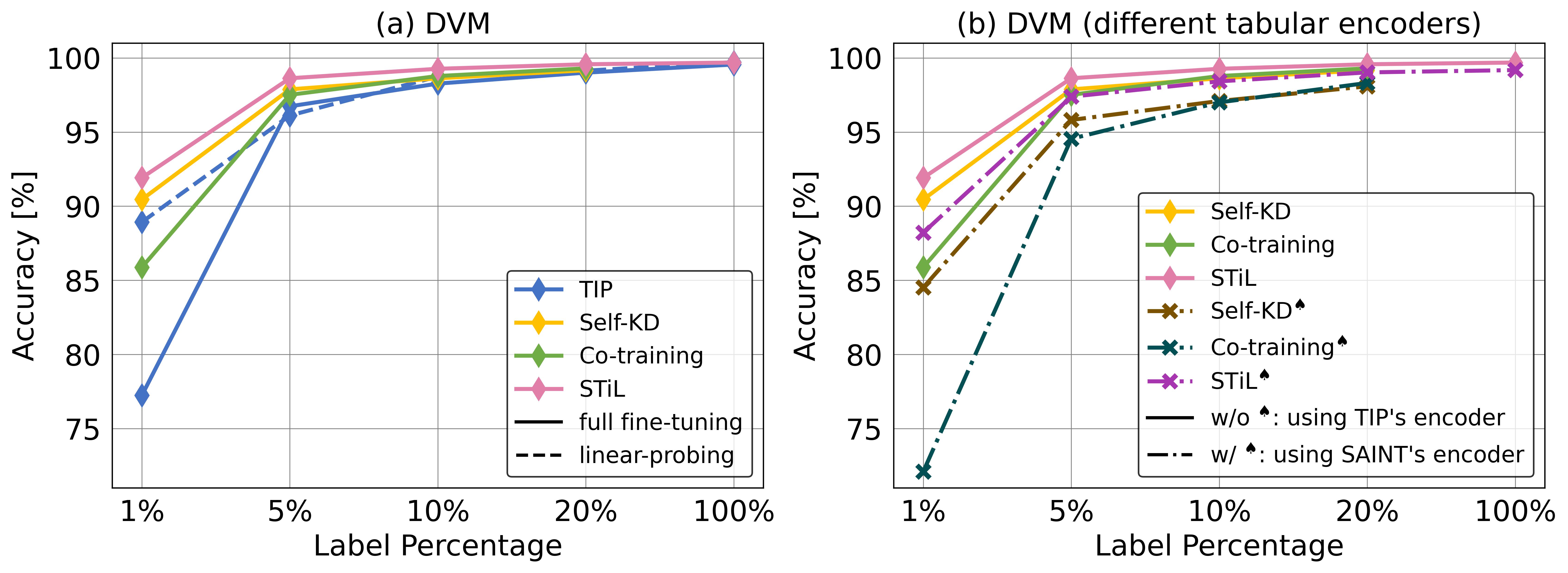}
  \caption{(a) Results comparing SSL and SemiSL multimodal SOTAs with \modelname{} using a finer grid of label percentage. (b) Results of SemiSL multimodal SOTAs and \modelname{} using different tabular encoders.}
   \label{fig:finer_low}
\end{figure*}

\begin{figure}
  \centering
  \includegraphics[width=0.80\linewidth]{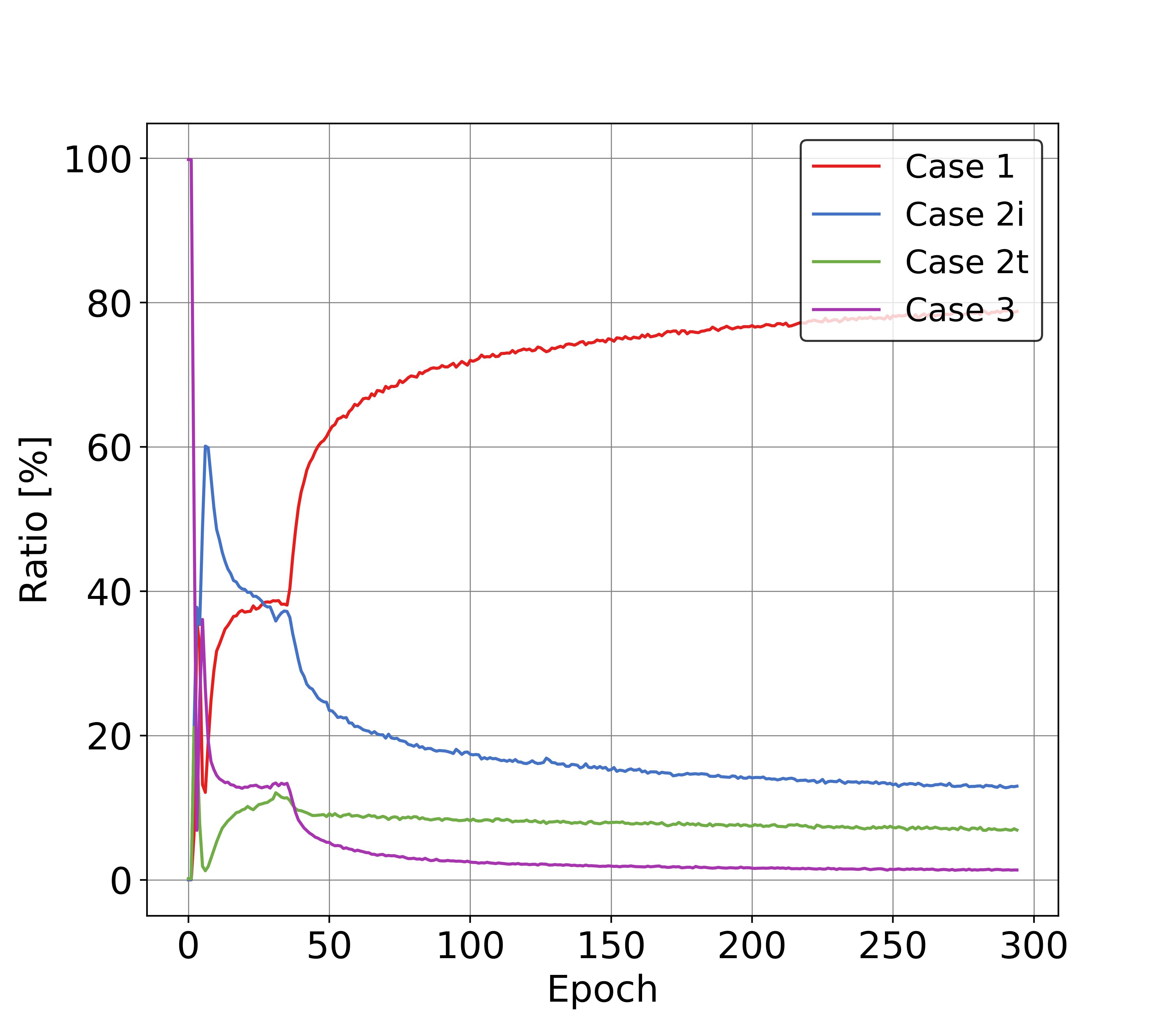}
  \caption{Sample ratios for each case in CGPL during training. The model is trained on 1\% labeled DVM.}
   \label{fig:CGPL_case}
\end{figure}

\begin{figure*}
  \centering
  \includegraphics[width=1\linewidth]{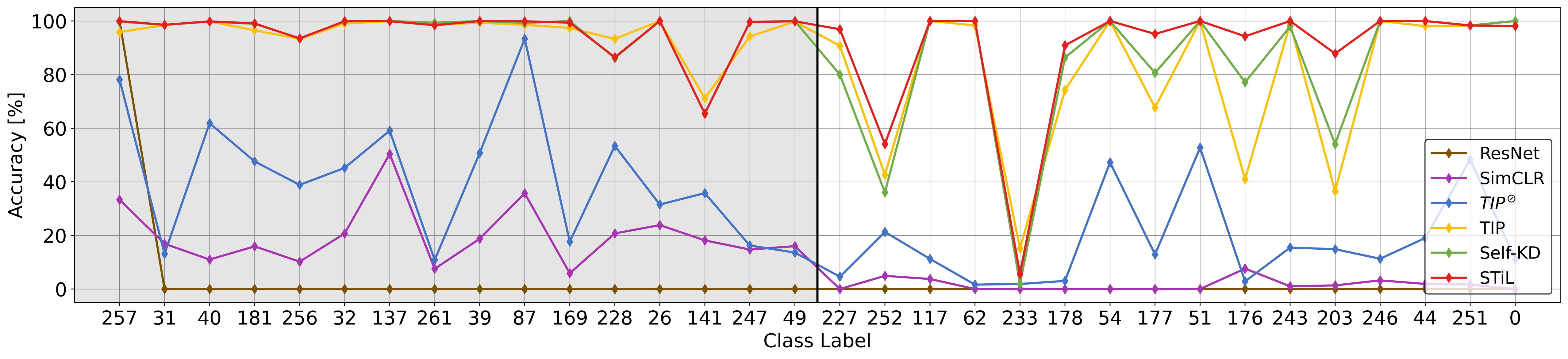}
  \caption{Class-wise accuracy comparing \modelname{} and other methods trained on 1\% labeled DVM. The top 16 majority classes are shown in the grey region, while the bottom 16 minority classes are shown in the white region. TIP$^{\oslash}$ represents TIP w/o SSL pre-training.}
   \label{fig:class_accuracy}
\end{figure*}

\begin{figure*}
  \centering
  \includegraphics[width=0.9\linewidth]{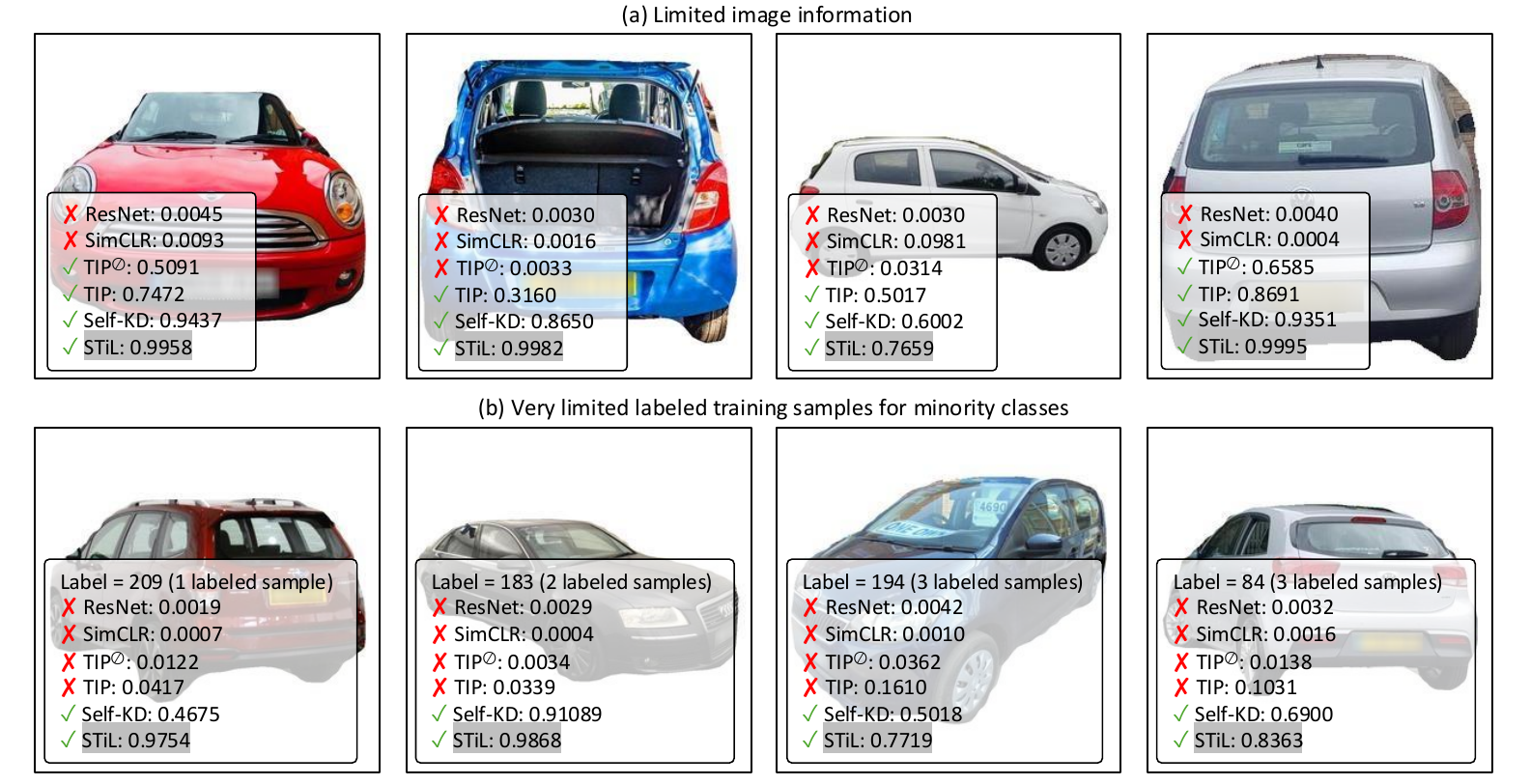}
  \caption{DVM car visualization of challenging samples and the ground-truth class predictions for \modelname{} and other models trained on 1\% labeled DVM. (a) Samples with limited image information, where the views of cars are restricted due to shooting angles (compared to the samples shown in (b)); (b) Samples from minority classes. \textcolor{customred}{$\boldsymbol{\times}$} indicates the model predicts a wrong class, while \textcolor{customgreen}{$\boldsymbol{\surd}$} indicates the model predicts the correct class.}
   \label{fig:case_study}
\end{figure*}

\noindent\textbf{SimMatch~\cite{zheng2022simmatch}:} This algorithm applies strong-to-weak consistency regularization at both the semantic and instance levels. It encourages different augmented views of the same instance to have the same class prediction and maintain similar similarity relationships with respect to other instances. Following the original paper, we set the weight factors for the unlabeled classification loss and the instance consistency regularization loss, \ie, $\lambda_u$ and $\lambda_{in}$, to 10 and 5, respectively. The smoothness parameter $\alpha$ was set to 0.9, the temperature parameter to 0.1, and the EMA momentum to 0.996. The batch sizes for labeled and unlabeled data were the same as those used in \modelname{}. Based on validation performance, we set the threshold in strong-to-weak consistency regularization to 0.8 for DVM and to 0.6 for CAD and Infarction. The starting pseudo-labeling epoch was 10 for DVM and 8 for CAD and Infarction.

\noindent\textbf{SimMatch$^{M}$:} This model is an adaptation of SimMatch to the multimodal image-tabular setting. The hyper-parameters were the same as those used in SimMatch. According to validation performance, we set the threshold for strong-to-weak consistency regularization $\tau$ to 0.9 for DVM and to 0.85 for CAD and Infarction.

\noindent\textbf{FreeMatch~\cite{wang2022freematch}:} This approach is also based on strong-to-weak consistency regularization and focuses on effectively leveraging unlabeled data. It adjusts the confident threshold in a self-adaptive manner according to the model's learning progress. Following the original paper, we set the weight factors for unlabeled classification loss and self-adaptive fairness loss, \ie, $w_u$ and $w_f$, to 1 and 0.001, respectively. The temperature parameter was set to 0.1, and the EMA momentum to 0.996. The batch sizes for the labeled and unlabeled data were the same as those used in \modelname{}. 

\noindent\textbf{FreeMatch$^M$:} This model is an extension of FreeMatch to the multimodal image-tabular setting. Its hyper-parameters were the same as those used in FreeMatch. 

\noindent\textbf{Co-training~\cite{blum1998combining}:} We adapt this co-pseudo-labeling based method to the multimodal image-tabular domain. The predictions from the image classifier serve as the pseudo-labels for the tabular classifier, and vice versa. Then a multimodal classifier trained on labeled data is used for classification. The weight factors for the labeled and unlabeled classification losses, \ie, $\alpha$ and $\lambda_u$, as well as the EMA momentum, were the same as those used in \modelname{}.

\noindent\textbf{MMatch~\cite{wang2022mmatch}:} In MMatch, predictions from a multimodal classifier are used as pseudo-labels for training unimodal classifiers. In addition, similar to CoMatch, MMatch refines the pseudo-labels by aggregating label information from nearby samples in the embedding space. Following the original paper, we set the smoothness parameter to 0.9, and the embedding memory bank size to 640. Based on validation performance, the weight factor for the unlabeled classification loss was set to 0.2.

\noindent\textbf{Self-KD~\cite{wang2024knowledge}:} This method is based on co-pseudo-labeling and cross-modal consistency regularization. In Self-KD, a multimodal classifier serves as the teacher for unimodal classifiers, transferring knowledge to them through pseudo-labeling. Meanwhile, the average ensemble of unimodal classifiers is used as the pseudo-label for training the multimodal classifier. Following the original paper, we set the weight factors for the knowledge distillation loss, the contrastive loss, and the L1-norm regularization term, \ie, $\gamma$, $\delta$, and $\eta$, to 0.6, 1, and 0.1, respectively. 

The learning rate and the number of parameters for each algorithm are summarized in~\cref{tab:learning_rate}. We used the Adam optimizer~\cite{kingma2014adam} without weight decay and deployed all models on 2 A5000 GPUs. To mitigate overfitting, similar to~\cite{du2024tip,hager2023best}, we employed an early stopping strategy in Pytorch Lightning, with a minimal delta (divergence threshold) of 0.0001, a maximal
number of epochs of 500, and a patience (stopping threshold) of 100 epochs. We ensured that all methods had converged under this training configuration. 

\section{Additional Experiment}

\noindent \textbf{Experiments with a Finer Grid of Label Percentage:} In Tab. 2 and Tab. 3 of the manuscript, we compared \modelname{} with SOTA SemiSL methods using experiments with 1\% and 10\% labeled samples. To provide a more detailed analysis, we further conducted experiments on DVM with additional label percentages of 5\%, 20\%, and 100\%, as shown in~\cref{fig:finer_low}(a). The results demonstrate that \modelname{} consistently outperforms SOTA SSL/SemiSL methods across different label percentages.

\noindent \textbf{Applicability to Different Tabular Encoders:} To demonstrate the general applicability of \modelname{}, we evaluated its performance with different tabular encoders. Specifically, we replaced TIP's pre-trained tabular encoder with SAINT~\cite{somepalli2021saint}'s pre-trained tabular encoder. As shown in~\cref{fig:finer_low}(b), all SemiSL approaches exhibit performance drops when using SAINT's encoder, indicating that TIP is a more powerful tabular encoder than SAINT, as also noted in TIP's paper~\cite{du2024tip}. However, while Self-KD and Co-training experience a significant performance decrease, \modelname{} remains more stable and continues to achieve the best performance, demonstrating its robustness across different tabular encoders.

\noindent \textbf{Sample Ratios for Different Cases in CGPL:} As mentioned in Sec 3.3 of the manuscript, CGPL categorizes samples into 4 cases based on classifier consensus: (1) Case 1: all classifiers agree; (2) Case 2i: $f^m$ and $f^i$ agree; (3) Case 2t: $f^m$ and $f^t$ agree; and (4) Case 3: none of the above. To assess the efficacy of CGPL, we visualize the changes in the ratios of the samples belonging to each case during training. As shown in~\cref{fig:CGPL_case}, the sample ratios for both case 2i and case 2t initially increase during the initial training stage but later decrease and stabilize at a lower bound. On the other hand, the sample ratio of Case 1 gradually increases and approaches an upper bound. These observations demonstrate that: (1) CGPL facilitates collaboration among classifiers, enabling them to learn from each other and improving classifier agreement; (2) due to the \emph{Information Modality Gap}, unimodal classifiers, which rely solely on a single modality, lack comprehensive task knowledge and fail to align with the multimodal classifier on certain challenging multimodal cases; and (3) CGPL effectively generates pseudo-labels through classifiers' consensus collaboration while allowing classifier diversity, which helps reduce the risk of classifier collusion.

\noindent \textbf{Class-wise Accuracy in DVM:} DVM has 283 classes, each with a varying number of labeled training samples. To investigate the impact of imbalanced data on \modelname{} and other comparing algorithms, we visualize their class-wise accuracy for both majority classes (those with more training samples) and minority classes (those with fewer training samples). Specifically, we ranked the classes based on their number of labeled training samples and displayed the class-wise accuracy for the top 16 majority classes and the bottom 16 minority classes. As shown in~\cref{fig:class_accuracy}, supervised methods exhibit low accuracy across different classes, indicating their limited capacity when trained with a few labeled data. Though TIP, the SSL pre-training framework, performs well on majority classes, its accuracy significantly decreases on some minority classes. This suggests that relying solely on a small amount of labeled data during fine-tuning is ineffective, especially for minority classes. In contrast, \modelname{} mitigates these issues by leveraging labeled and unlabeled data jointly, achieving overall better results. In addition, we observe that all models perform poorly on class 233, which can be attributed to the very limited labeled data (only 1 training sample) and the inherent difficulty in classifying this class.

\noindent \textbf{Case Study:} We visualize several challenging examples where \modelname{} outperforms previous SOTAs. The results show that (1) a single image modality is insufficient to solve the classification task (the failure of ResNet and SimCLR in \cref{fig:case_study}(a)) and (2) minority classes with very limited labeled training samples pose challenges for SSL algorithms (the failure of TIP in \cref{fig:case_study}(b)). However, \modelname{} enables the model to comprehensively explore task-relevant information from both labeled and unlabeled data, leading to improved performance on these challenging samples.




\end{document}